\documentclass[10pt,a4paper,twocolumn]{article}

\input{pisika.dat}

\begin{document}

\providecommand{\ShortAuthorList}[0]{Boyi Jin} 
\title{An intelligent algorithmic trading based on a risk-return reinforcement learning algorithm}
\author[1]{Boyi Jin\thanks{Corresponding author: 20110983@sdufe.edu.cn}}

\affil[1]{Department of Insurance, Shandong University of Finance and Economic,40 shungeng Road,shizhong District, Jinan, 250002, China}


\begin{abstract}
\noindent
It is a challenging problem to automatically generate trading signals based on historical transaction information and the financial status of assets. This scientific paper proposes a novel portfolio optimization model using an improved deep reinforcement learning algorithm. The optimization model's objective function considers both a financial portfolio's risks and returns. The proposed algorithm is based on actor-critic architecture, in which the main task of the critic network is to learn the distribution of cumulative return using quantile regression, and the actor network outputs the optimal portfolio weight by maximizing the objective function mentioned above. Meanwhile, we exploit a linear transformation function to realize asset short selling. Finally, A multi-process method, called Ape-x, is used to accelerate the speed of deep reinforcement learning training. To validate our proposed approach, we conduct backtesting for two representative portfolios and observe that the proposed model in this work is superior to the benchmark strategies.\par

\DOI{} 
\end{abstract}

\maketitle
\thispagestyle{titlestyle}

\section{Introduction}\label{sec:intro}
Algorithmic trading, which has been widely used in the financial market, is a technique that uses a computer program to automate buying and selling stocks, options, futures, cryptocurrency, etc. Institutional investors such as a pension, mutual, and hedge funds usually use algorithmic trading to find the most favorable execution price, reduce the impact of severe market fluctuations and improve execution efficiency. Algorithmic trading has a far-reaching effect on the overall efficiency and microstructure of the capital market. Therefore, asset pricing, portfolio investment, and risk measurement may undergo revolutionary changes.\par
        Classical algorithmic strategies includes arrival price  strategy\cite{perold1988implementation,almgren2007adaptive}; volume weighted average price  strategy\cite{berkowitz1988total,madhavan2002first}; time-weighted average price strategy\cite{kolm2011algorithmic}, implementation shortfall strategy\cite{kritzman2006optimizers,hendershott2013algorithmic}; guerrilla strategy\cite{eriksson2012ucd},etc. In recent years, with the rapid development of artificial intelligence(AI), more researchers have begun utilizing deep reinforcement learning to implement algorithmic trading. Deep reinforcement learning mainly uses the excellent feature representation ability of the deep neural network to fit the state, action, value, and other reinforcement learning functions to maximize portfolio return. \par
        Currently, researchers exploited different reinforcement learning algorithms to study financial trading problems, including the actor-only method(also called the value-based method), the critic-only method(also called the policy-based method), and the actor-critic method. Nevertheless, we believe these studies are not yet mature in practical application for the following reasons. First, most studies maximize the average cumulative rewards, which is also the expectation of cumulative rewards, of a single asset or a portfolio by training value function(such as policy gradient) or action-value function(such as Q-learning). However, these algorithms do not consider the risk and are only suitable for investors as a neutral risk type. A large number of studies\cite{jondeau2003testing,tsiakas2006periodic,franke2009applying} have proved that the stock return has fat tail characteristics, so the low-probability tail risk needs to be highly concerned.\par
Additionally, short selling is allowed in many stock markets. Short selling can not only make a profit in a bear market but also reduce the speculation and volatility of the stock market. Although \cite{park2020intelligent} and \cite{almahdi2019constrained}  have considered short selling, these studies focus on single asset trading. Most traders generally hold multiple securities. Unfortunately, most existing studies do not consider the short-selling problem in this common case.\par
Finally, the interaction between agent and environment is often very time-consuming. However, profit opportunities are fleeting in the stock market. Therefore, how to improve the training speed is also worth further study. \par
The main contribution of this paper is as follows:\par
Firstly, this study proposes a new algorithm for the actor-critic framework, called risk-return reinforcement learning(R3L). Motivated by the modern portfolio theory proposed by \cite{markowitz1952harry}, we construct an optimization model based on risk and return. The objective function of the optimization model is the weighted sum of mean and value at risk(VaR) of portfolio cumulative return. The main goal of the actor network is to learn the portfolio weights by maximizing the objective function mentioned above. The primary purpose of the critic network is to learn the cumulative portfolio return distribution. Inspired by distributional reinforcement learning proposed by \cite{dabney2018distributional}, we use quantile regression to estimate the parameters of the critic network. \par
Secondly, similar to the approach proposed by \cite{betancourt2021deep}, we use a soft-max function to transform the output of the actor network into a new variable to meet the self-financing constraint of the portfolio. Then, a linear transformation is implemented to convert the variable mentioned above into the final output, i.e., portfolio weight. This transformation still meets the self-financing constraint. Meanwhile, it can realize short selling, control the scale of short selling and prevent radical investment strategies.\par
Thirdly, we leverage a multi-process algorithm, called Ape-x algorithm proposed by \cite{horgan2018distributed}, to speed up the training. Ape-x algorithm separates data collection from strategy learning, use multiple parallel agents to collect experience data, share a large experience data buffer, and sends it to learners for learning. \par
The rest of this paper is organized as follows. In Section \ref{section2}, we review the related literature and present the differences between our study and previous studies. Section \ref{section3} describes the definition of our problem, and Section \ref{section4} introduces  our proposed R3L algorithm. In Section \ref{section5}, we detail the setups of our experiments. In Section \ref{section6}, we provide the experiment result. Finally, Section \ref{section7} presents the conclusions and directions of future work.
\section[2]{Related works}\label{section2}\par
In this section, we review literature works regarding deep reinforcement learning in financial trading. As mentioned in the introduction, the algorithms used in financial trading mainly include the critic-only method, the actor-only method, and the actor-critic method. Then, we will elaborate on the application of the three methods.
\subsection[2.1]{The critic-only method}
The critic-only method is also called the value-based method. DQN(Deep Q-Learning Network) is a  well-known value-based deep reinforcement learning algorithm, the core of which is to use a neural network to replace the q-label, i. e. the action value function. The input of the neural network is state information, and the output is the value of each action. Therefore, The DQN algorithm can be used to solve the problem of continuous state space and discrete action space but can not solve the problem of continuous action space.\par
DQN and its improved algorithms have received extensive attention in the research of financial trading.
\cite{alimoradi2018league} proposed the league championship algorithm (LCA)  that can extract and save various stock trading rules according to different stock market conditions. Every agent in LCA is a sports team member and can learn from the weaknesses and strengths of others to improve their performance. They proved that their algorithm outperforms Buy-and-Hold and GNP-RA strategies.
\cite{bu2018learning}trained an LSTM network using double Q-learning \cite{van2016deep} and obtained positive rewards in a cryptocurrency market with a decreasing tendency.
\cite{chen2019application} proposed a Deep Recurrent Q- Network(DRQN), which can process sequences better. They found that the proposed algorithm outperformed the buy and hold strategy in the S\&P500 ETF history dataset.
\cite{huang2018financial} also adopted a DRQN-based algorithm, which was applied to the foreign exchange market from 2012 to 2017, and developed a motion enhancement technology to mitigate the lack of exploration in Q-learning. Experimental results showed that the annual return of some currencies was 60\%, with an average of about 10\%.
\cite{lucarelli2020deep} proposed a novel deep Q-learning portfolio management strategy. The framework consists of two main components: a group of local agents responsible for trading a single asset and a global agent that rewards each local agent based on the same reward function. The results showed that the proposed approach was a promising method for dynamic portfolio optimization.
The deep Q-learning network proposed by \cite{jeong2019improving}  is based on an improved DNN structure consisting of two branches, one of which learned action values while the other learned the number of shares to take to maximize the objective function.
\cite{zhang2020deep} used a LSTM based Q network to implement portfolio optimization. In addition, to consider the risk factors in portfolio management, the volatility of portfolio return was pulsed to the rewards function.
\cite{park2020intelligent} discretized the action space by fixing the trading amount, in which trader can buy or sell a specific asset for a fixed amount. Short selling is not allowed, fixed the trading amount for each asset,
and a mapping function is designed to transform the infeasible action into feasible action.
\subsection[2.2]{{Actor-only  method}}
 In the actor-only method, the action taken by the agent can be learned directly by a neural network. The advantage of this method is that it makes it available for continuous action space. \cite{deng2016deep} constructed a recursive deep neural network for environmental perception and recursive decision-making. Then, deep learning is used for feature extraction and combined with fuzzy learning to reduce the uncertainty of input data. Another novelty of this paper is a variant of Backpropagation through time(BTT) to handle the vanishing gradient problem. \cite{wang2016dueling} used Policy Gradient(PG) approach for financial trading. Its main contribution is to analyze the advantages of the LSTM network structure over the fully connected network and to compare the impact of some combination of technical indicators on revenue. The drawback of the Actor-only method is that, due to the policy strategy, the number of interactions between agents and the environment is greatly increased compared with a value-based method. As a result, the training is very time-consuming.
 \subsection[2.3]{Actor-Critic  method}
The actor-critic method contains two parts: an actor, which selects action based on probability or certainty, and a critic, which evaluates the score of the action taken by the actor. The actor-critic method combines the advantages of the value-based and policy-based methods, which can not only deal with continuous and discrete problems but also carry out the one-step update to improve learning efficiency.
\cite{wu2020adaptive} exploited Gated Recurrent Unit(GRU) to extract financial features and proposed a critic-only method called Gated Deep Q-learning trading strategy(GDQN) and Actor-Critic method called Gated Deterministic Policy Gradient trading strategy(GDPG). Experimental results show that GDQN and GDPG outperformed the Turtle trading strategy\cite{vezeris2019adturtle} and DRL trading strategy\cite{deng2016deep}, and the performance of GDPG is more stable than the GDQN in the ever-evolving stock market.
\cite{liang2018adversarial} adopted three versions of the RL algorithm based on Deep Deterministic Policy Gradient (DDPG), Proximal Policy Optimization (PPO), and Policy Gradient (PG) for portfolio management. A so-called Adversarial Training was proposed to reach a robust result and to consider possible risks in optimizing the portfolio more carefully.
\cite{betancourt2021deep} proposed a novel reinforcement learning algorithm for portfolio optimization using Proximal Policy Optimization (PPO). An essential characteristic of the proposed model is that the number of assets is dynamic. Experiments on a cryptocurrency market showed that the proposed algorithm outperformed three state-of-the-art algorithms presented by \cite{bu2018learning,jiang2017deep,pendharkar2018trading}.
\cite{schnaubelt2022deep} compared the performance of deep double Q-learning and proximal policy optimization (PPO) to several benchmark execution policies and found that PPO realizes the lowest total implementation short-fall across exchanges and currency pairs. \par
To sum up, these studies used various RL algorithms for financial trading problems in a different settings. The existing literature also has the following problem.
First, most literature determines the optimal action(portfolio decision) based on the mean of the portfolio cumulative returns(for example, Q- value) and pays less attention to the risk. Although \cite{lucarelli2020deep,wu2020adaptive,betancourt2021deep} consider the risk factor in the reward function,  these authors can not measure the risk of portfolio cumulative return.
Second, short selling is allowed in many strategies\cite{bertoluzzo2012testing,deng2016deep,almahdi2019constrained,jeong2019improving}, but the above strategies consider trading for only one asset, which is often inconsistent with the facts; most of the investors often hold multiple assets to diversify risk. Although some strategies\cite{jiang2017deep,almahdi2019constrained,park2020intelligent} consider trading various assets, short selling was not allowed, which means that investors will not be able to make profits in the bear market.
Third, the interaction between agent and environment is very time-consuming in training. Most existing literature does not consider how to improve the training speed.
Thus, the algorithm proposed in this paper can effectively overcome the shortcomings mentioned above.
\section[3]{Preliminaries}\label{section3}
In this section, we introduce some preliminaries about MDP and elaborate state space, action space, and rewards function of the R3L algorithm.
\subsection[3.1]{Markov decision process}
Markov decision process (MDP) is a mathematical model for sequential decision problems, which is used to simulate the random strategies and rewards that agents can realize in the environment where the system state has Markov properties. Portfolio optimization is a typical sequential decision problem. Investors dynamically adjust portfolio weights according to market information. Thus, we can consider the portfolio allocation problems as a Markov Decision Process (MDP). If a state transition conforms to Markov, the next state only depends on its current state and has nothing to do with the state before its current state. A finite MDP (as considered here) is a four tuple, denoted by  $(S,A,P_a(s,s'),R_a(s,s'))$.\par
where:\par
$\bullet$ S is a finite set of states.\par
$\bullet$ A is a finite set of actions (and As is the finite set of actions available from state s).\par
$\bullet$  $R_a(s,s')$ is reward function.\par
$\bullet$ $P_a(s,s')=Pr(s_{t+1}=s'|s_{t}=s,a_{t}=a)$ is the state transition probability.\par
The interaction between an agent and financial environment will produce a trajectory $\tau=\{s_0,a_0,R_0,s_1,a_1,R_1,...\}$. $G_t$ is the discounted cumulative reward, which the agent can obtain at time t expressed as follows:
\begin{equation}\label{equation:5}
  G_{t}=\sum_{i=t}^{T}\gamma^{i-t}R(s_i,s_{i+1})
\end{equation}
where $\gamma\in [0,1]$ is the discounted rate.\par
To learn the optimal strategy, we use the value function. There are two types of value functions in reinforcement learning: state value function, denoted by $V(s)$, and action-value function, denoted by $Q(s,a)$. The state value function, shown in Eq.(\ref{Vs}), represents the expectation of cumulative rewards from a certain state s.
\begin{equation}\label{Vs}
  V(s)=E[G_t|s_t=s]
\end{equation}\par
The action value function, given in Eq.(\ref{Q}),  is the expected return obtained after the agent executes an action in the current state s.\par
\begin{equation}\label{Q}
  Q(s,a)=E[G_t|s_t=s,a_t=a]
\end{equation}
\subsection[3.2]{State space}\par
An important and difficult problem in algorithmic trading is the low observability of the market environment. The information available to investors is extremely limited compared to the complex market environment. Therefore, how to deal with this limited information is very important.
In this paper, the state at the period t, denoted by $s_{t}$, consists of three types of variables: historical data for the assets($K_t$), portfolio weights($w_t$) and trading time step(t). The input of the actor network is historical data, and the input of the critic network is the concatenation of historical data, portfolio weight, and the time index. The historical data of the selected portfolio consists of raw data and technical indicators. The raw data includes the price open, close, high, low, and volume(OCHLV). Technical indicators consist of a list of candlestick patterns, including bearish, bullish, significant, hammer, inverse hammer, bullish engulfing, piercing line, morning star, bullish harami, hanging man, shooting star, bearish engulfing, evening star, three black crows, dark cloud cover, bearish harami, doji, spinning top. See \cite{taghian2022learning} for details. All historical data can be represented by a tensor with dimension $n \times g \times h$, where n is the number of risky assets, g is the number of features per asset, h denotes a window size which is the number of latest feature value the agent can observe. Portfolio weight, also called portfolio vector, is the percentage of a total portfolio represented by a single asset. The rationality of adding a time index to the critic network is that it reflects the time value of money. \par
\begin{equation}\label{states}
\begin{split}
&s_t=(K_t,w_t,t)\\
&w_t=(w_{1t},w_{2t},...w_{n,t})\\
&K_t=(K_{1t},K_{2t},...K_{n,t})
\end{split}
\end{equation}\par

\subsection[3.4]{Action space}
A problem in algorithmic trading is the low observability of the market environment. Compared with the complexity of the market environment, the information available to investors is extremely limited. Therefore, how to deal with little details is very important.
At each time step t, the agent executes a trading action resulting from the actor network. Specifically, the agent needs to redetermine the optimal portfolio weight according to the updated state at the end of each period. In addition, to realize short selling, the actor network needs to be modified as follows: Firstly, to meet self-financing conditions, we use the softmax function to transform the output of the actor network into a new variable, denoted by $SA=\{SA_{1},SA_{2},...SA_{n}\}$. This transformation function is given in Eq.(\ref{transformation}).
\begin{equation}\label{transformation}
  SA_{i}=\frac{exp^{A_{i}}}{\sum_{i=0}^{n-1}exp^{A_{i}}},\quad \quad i=1,2,...n
\end{equation}\par
where  $A=\{A_{1},A_{2},...A_{n}\}$ is the initial output of the actor network.\par
The default of this transformation is that the portfolio weight of each asset is greater than zero, which makes short selling impossible. Short selling occurs when an investor borrows a security and sells it on the open market, planning to repurchase it later for less money. It allows investors to benefit not only from a bear market but also to use capital proceeds to overweight the portfolio's long-only component of the portfolio. Different from the existing literature, we realize short selling of assets through the following transformation in Eq.(\ref{liner_transformation}), in which $w_{i}(i=1,2,\cdot\cdot\cdot n)$ represents the proportion of asset i after transformation and $\delta$(called Delta) represents adjustment parameter.
\begin{equation}\label{liner_transformation}
  w_{i}=SA_{i} \times\delta-\frac{\delta-1}{n},\quad \quad i=1,2,...n
\end{equation}\par
The self-financing constraint can be satisfied through the above linear transformation, and this transformation can also realize short selling. In particular, $\delta=1$ indicate  no change of portfolio weight, $\delta=0$ indicate  that the portfolio weight of each asset is equal to $1/n$. Furthermore, in this paper, our portfolio includes four risky assets and a risk-free asset(n=5); the adjustment parameter is set to 3($\delta=3$), so the portfolio weight of a single asset ranges from -40\% to 260\%, which means the maximum proportion of short selling in total assets is 160\%.
\subsection[3.4]{rewards}
The reward reflects the performance of an agent's action. Let $asset_{t}$ denote portfolio value at the end of period $t$. The reward, denoted by $reward_{t}$, is defined as portfolio return in period t, computed as the following equation:
\begin{equation}\label{reward}
  reward_{t}=\frac{asset_{t}}{asset_{t-1}}-1
\end{equation}\par
If we do not take  into account transaction cost, portfolio value at the end of period $t$ equals portfolio value at the end of  period $t-1$ plus portfolio return, expressed as follows:
\begin{equation}\label{asset(t)}
  asset_{t}= asset_{t-1}+asset_{t-1}\times\sum_{i=1}^{n}w_{i,t-1}\times R_{i,t}
\end{equation}\par
If we do not take into account transaction cost, portfolio value at the end of period $t$ equals portfolio value at the end of  period $t-1$ plus portfolio return, expressed as follows
\begin{equation}\label{equationio}
  reward_{t}=\sum_{i=1}^{n}w_{i,t-1}R_{i,t}
\end{equation}\par
If we consider transaction cost, portfolio value at the end of period $t$ equals  $asset_{t-1}$ plus portfolio return  minus transaction costs, computed as follows:\par
\begin{equation}\label{equationwe}
\begin{split}
  asset_{t}=& asset_{t-1}+asset_{t-1}\times\sum_{i=1}^{n}w_{i,t-1}\times R_{i,t}-\\
  &c1\times \sum_{i=1}^{n}sell_{i,t}\times sellin_{i,t} \times price_{i,t}-\\
  &c2\times \sum_{i=1}^{n}buy_{i,t}\times buyin_{i,t}\times price_{i,t}\\
\end{split}
\end{equation}
where:\par
 $\bullet$ c1 and c2 is the transaction cost for selling and buying.\par
 $\bullet$ $sell_{i,t}$ and $buy_{i,t}$ are dummy variables, indicating whether or not to buy or sell asset $i$ at the end of period t.\par
 $\bullet$ $buyin_{i,t}$ and $sellin_{i,t}$ denote the trade size of asset i at the end of period t, which should be greater than or equal to zero. \par
 In addition, the share of asset i held by an investor at the end of period t, represented by $asset_{t}\times w_{i,t}$, satisfies the following equation:\par
\begin{equation}\label{equationhj}
\begin{split}
  asset_{t}\times w_{i,t}= &asset_{t-1}\times w_{i,t-1}\times (1+R_{i,t})+\\
  &buy_{i,t}\times buyin_{i,t}\times price_{i,t}-\\
  &sell_{i,t}\times sellin_{i,t}\times price_{i,t}
\end{split}
\end{equation}\par
 Obviously, $asset_{t}\times w_{i,t}$  equals to the cumulative value of $asset_{t-1}\times w_{i,t-1}$ in period t plus the shares bought($buy_{i,t}\times buyin_{i,t}\times price_{i,t}$) minus the shares sold($sell_{i,t}\times sellin_{i,t}\times price_{i,t}$). Since it does not make sense to buy and sell an asset simultaneously, which increases transaction costs, one of $sell_{i,t}$ and  $buy_{i,t}$ must be zero.\par
If $asset_{t-1}$ , $w_{i,t-1}$ and $w_{i,t}$ are given, we could adjust  trading strategies to maximize the portfolio value $asset_{t}$. This optimization problem can be formulated as a nonlinear programming, the objective function of which is to maximize $asset_{t}$, and the decision variables are $sell_{i,t}$ , $buy_{i,t}$, $buyin_{i,t}$ and $sellin_{i,t}$. The nonlinear programming model can be expressed as follows:
\begin{equation}\label{optimization}
\begin{split}
&\mathbf{Maximize:} \quad asset_{t}\\
&\mathbf{Subject\quad to}:\\
  &\quad  asset_{t}\times w_{i,t}= asset_{t-1}\times w_{i,t-1}\times(1+R_{i,t-1})-\\
  &\quad \quad \quad \quad \quad \quad sell_{i,t}\times sellin_{i,t}\times price_{i,t}+\\
  &  \quad \quad \quad \quad \quad \quad buy_{i,t}\times buyin_{i,t}\times price_{i,t}\\
  &\quad asset_{t}= asset_{t-1}+asset_{t-1}\times\sum_{i=1}^{n}(w_{i,t-1}\times R_{i,t-1})-\\
  & \quad \quad \quad \quad c1\times \sum_{i=1}^{n}sell_{i,t}\times sellin_{i,t} \times price_{i,t}-\\
  &\quad \quad \quad \quad c2\times \sum_{i=1}^{n}buy_{i,t}\times buyin_{i,t}\times price_{i,t}\\
  &\quad sell_{i,t}\times buy_{i,t}=0\\
&\quad sell_{i,t}+buy_{i,t}=1\\
&\quad sellin_{i,t}\geq0\\
&\quad buyin_{i,t}\geq0
\end{split}
\end{equation}\par
\section[4]{ methodology}\label{section4}
In the first part of this section, we describe the risk-return reinforcement learning(R3L) algorithm in detail. The second part of this paper introduces the architecture of the neural network. The third part of this section is Ape-x.
\subsection[4.1]{Proposed algorithm}
The algorithm adopted by this paper is based on actor-critic architecture, which includes an actor network and a critic network. In addition, each network has its corresponding target network, so the algorithm includes four networks, namely the actor network, denoted by $ \mu_{\varphi}(s) $, and the critical network, denoted by $ K_{\omega}(s,a)$, the target actor network, denoted by $\mu'_{\varphi'} (s)$ and the target critical network, denoted by $ K' _{\omega'}(s,a)$ . In classical actor-critic architectures such as A3C, TD3, and DDPG, the actor network updates $\varphi$ by maximizing the expectation of cumulative rewards, and the critic network updates $\omega$ by minimizing the error between the evaluation value and the target value. The algorithm proposed in this paper is different from the above algorithms. Inspired by distributional reinforcement learning(DRL) initially proposed by\cite{bellemare2017distributional}, we estimate the distribution, rather than the expectation, of cumulative rewards by the critic network.\par
Distributional reinforcement learning is a new kind of reinforcement learning algorithm, mainly learning the distribution of cumulative rewards. The distributional Bellman operator($\tau$) is shown in Eq.(\ref{bellman operator}).
\begin{equation}\label{bellman operator}
\tau  Z(s,a) \overset{D}{=}R(s,a)+\gamma Z(s',a')
\end{equation}
where $ Z(s,a)$ represents the cumulative return obtained by taking action $a$ in state $s$, which is a random variable,  $R(s,a)$ is the rewards function. \par
Under Wasserstein metric, \cite{bellemare2017distributional} proved that the distributional Bellman operator is a $\gamma$-contract operator. The learning task of DRL is to make the distribution $Z(s,a)$ and the target distribution $R(s,a)+\gamma Z(s',a') $ as similar as possible. Following \cite{dabney2018distributional}, we utilize quantile regression to estimate network parameters. Quantile regression, projecting the update of distributional Bellman to the quantile distribution,  uses a parameterized quantile distribution to approximate the value distribution. Let $ [\theta_{1}, \theta_ {2}, \ldots, \theta_ {N}]$, which is the output of critic network, denotes the N quantiles of $Z(s,a)$. The target distribution is shown in Eq.(\ref{target distribution}), which can be regarded as ground truth.
\begin{equation}\label{target distribution}
  \tau \theta'_{j}=r+\gamma \theta'_{j}, \quad \forall j
\end{equation}
The loss function of critic network is defined in Eq.(\ref{loss function}).
\begin{equation}\label{loss function}
  L_{\omega}=\frac{1}{N}\sum_{i=1}^{N}\sum_{j=1}^{N}[\rho_{\hat{\tau}_i}(\tau\theta'_{j}-\theta_i)]
\end{equation}
where
\begin{equation}\label{90}
\rho_{\tau}=|\tau-\delta_{\mu<0}||u|=(\tau-\delta_{\mu<0})u
\end{equation}
Because $|u|$ is not differentiable at zero, we take the Huber loss function, given in Eq.(\ref{Huber}), instead of $|u|$.
\begin{equation}\label{Huber}
\Gamma_{\kappa}=\left\{
\begin{aligned}
 & \frac{1}{2}u^2  &if \quad |u|\leq \kappa \\
 & \kappa (|u|-\frac{1}{2}\kappa) &  otherwise \\
\end{aligned}
\right.
\end{equation}\par
Thus, we get a new loss function, also called the quantile  Huber loss function, expressed as follows:
\begin{equation}\label{Huber loss}
  \rho_{\tau}^{\kappa}=|\tau-\delta_{\mu<0}|\Gamma_{\kappa}
\end{equation}\par
Since portfolio weight is a continuous variable, we choose a deterministic policy to generate actions, given the Eq.(\ref{action}).
\begin{equation}\label{action}
  a=\mu_{\varphi}(s)
\end{equation}
As mentioned in the introduction section, the objective function of portfolio optimization is the weighted sum of the mean and VaR of portfolio cumulative return. If we take N uniform quantiles, the mean of portfolio cumulative return, denoted by $MR$, can be approximately equal to the average of quantiles of cumulative return, shown in  Eq.(\ref{MR}).\par
\begin{equation}\label{MR}
  MR=\frac{\sum_{i=1}^{N}\theta_i}{N}
\end{equation}\par
VaR is used to measure risk in this paper. VaR refers to the maximum portfolio loss for a given confidence level in a specific period. It can be described in Eq.(\ref{var definition}), in which $R_p$ and $\alpha$ denote portfolio cumulative return and confidence level respectively.
\begin{equation}\label{var definition}
  Prob(R_p \leq -VAR_{\alpha})=1-\alpha
\end{equation}\par
Because ${VaR}_{\alpha}$ is a quantile of portfolio cumulative return, in this paper, it can also be expressed as the following:
\begin{equation}\label{VAR}
 VaR_{\alpha}=-\theta_{N\times(1-\alpha)}
\end{equation}\par
Given the value of N and $\alpha$, we can easily get $VaR_{\alpha}$. For example, if $N=100, \alpha=95\%$, $VaR_{95\%}=-\theta_{5}$, if $N=200,\alpha=90\%$, $VaR_{90\%}=-\theta_{20}$.\par
According to classical modern portfolio theory, portfolio selection aims to construct an optimal portfolio model that maximizes expected returns under a given acceptable risk level($\varpi$). This optimal model is shown in the following:
\begin{equation}\label{objective1 function}
\begin{split}
Max &\quad   MR\\
   &s.t. VaR_{\alpha}=\varpi
\end{split}
\end{equation}\par
According to the above optimization model, different risk levels correspond to optimal weights. In other words, investors have various portfolio choices, which undoubtedly increases the difficulty of decision-making. To obtain a single optimal solution, The above programming problem can be transformed into a single objective programming problem through the following function:
\begin{equation}\label{objective function}
\begin{split}
U&= MR-\zeta\times VAR_{\alpha}\\
   &=\frac{\sum_{i=1}^{N}\theta_i}{N}+\zeta\times \theta_{N*(1-\alpha)}
\end{split}
\end{equation}
where $\zeta$ (called Zeta) denotes the risk attitude of the investor, the higher $\zeta$, the higher the risk aversion of investors, and the more conservative investment strategies will be adopted.\par
Exploration is crucial for agents, and deterministic strategies cannot explore, so we need to artificially add noise to the output actions. In DDPG algorithm, \cite{lillicrap2015continuous} uses Ornstein Uhlenbeck process(OU) as action noise. \cite{fujimoto2018addressing} found noise drawn from the OU process offered no performance benefits, so in this paper, following \cite{fujimoto2018addressing}, we add Gaussian noise, which is given in Eq.(\ref{noise}), to each action. The standard deviation of Gaussian noise decreases exponentially as the number of parameter updates continues.
\begin{equation}\label{noise}
\begin{split}
noise\_  output_{i}& =actor\_output_{i}+\epsilon\\
& \epsilon \sim clip(N(0,\sigma),-c,c)
\end{split}
\end{equation}
\begin{figure*}[t]
  \centering
  \setlength{\abovecaptionskip}{0.cm}
  \setlength{\belowcaptionskip}{-0.cm}
  \includegraphics[height=10cm,width=16cm]{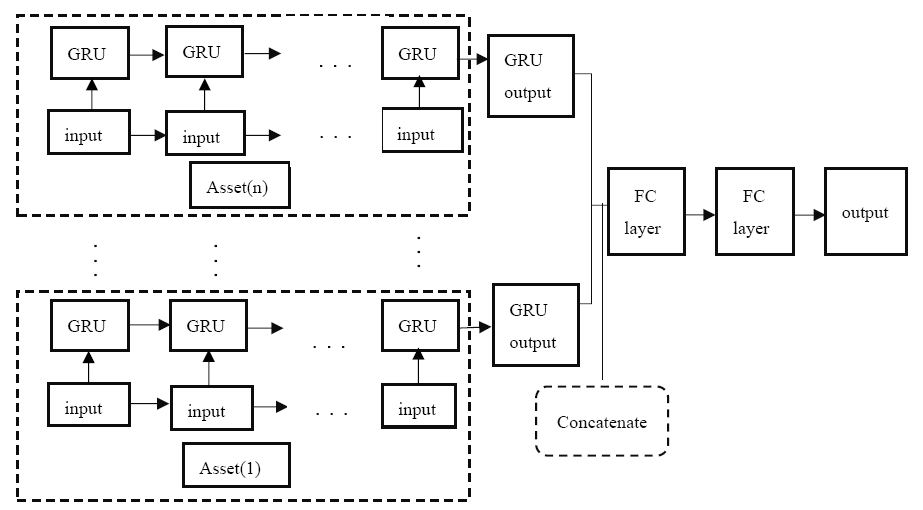}
  \caption{Actor network structure}
  \label{Actor}
\end{figure*}
\subsection[4.2]{Neural network}\label{section4.2}
\begin{figure*}[t]
  \centering
  \setlength{\abovecaptionskip}{-0.5cm}
  \setlength{\belowcaptionskip}{-0.5cm}
  \includegraphics[height=10cm,width=16cm]{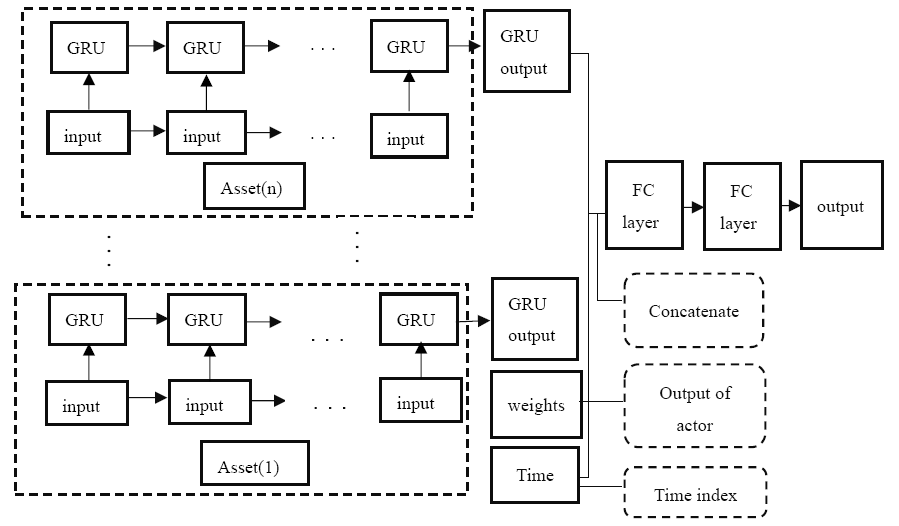}
  \caption{Critic network structure}
  \label{critic}
\end{figure*}

We need to model the neural network structure to explore functional patterns and extract informative features. Based on the time series nature of stock data, the gated recurrent unit(GRU) is utilized to construct an informative feature representation. The GRU has two gates, i.e., a reset gate and an update gate. These two gates determine which information can ultimately be used as the output of the gating recurrent unit. They are unique in that they can preserve information in long-term sequences and will not be cleared over time or removed because they are unrelated to prediction. Intuitively, the reset gate determines how to combine the new input information with the previous memory. The update gate defines the amount of prior memory saved to the current time step. The actor and critic network structure is shown in Fig.\ref{Actor} and Fig.\ref{critic}. Since GRU, LSTM, and RNN have similar structural features, we also used LSTM and RNN networks in the sensitivity analysis.

In addition, Several authors have leveraged Convolutional Neural Networks(CNN) to perform financial trading strategy\cite{carta2021multi,barra2020deep}, treating the stock trading problem as a computer vision application by using time-series representative images as input. The convolutional neural network has unique advantages in speech recognition and image processing with its particular structure of local weight sharing. Its design is closer to the actual biological neural network. Weight sharing reduces the complexity of the network, especially the feature that the images of multi-dimensional input vectors can be directly input into the network, which avoids the complexity of data reconstruction in the feature extraction and classification process. Because of CNN's robust feature representation ability, we studied the impact of CNN on the performance of the proposed algorithm in the sensitivity analysis.
\subsection[4.3]{Distributed framework}
In this paper, to make the training result robust, the actor must interact with the environment at different epochs(or iterations) several times, which is very time-consuming. We utilize  Ape-X architecture, proposed by \cite{horgan2018distributed}, to speed up the training procedure. This algorithm decouples acting from learning and decomposes the learning process into three parts. In the first part, there are multiple actors. Each actor interacts with its respective environment based on the shared neural network, accumulating experience and putting it into the shared experience replay memory. We refer to this part, running on CPUs, as acting. In the second part, the (single) learner samples data from the replay memory and then updates the neural network. We refer to this part as the learning part running on a GPU. The third part is mainly responsible for data transmission. We refer to this part, running on CPUs, as communication.\par
\begin{figure*}[t]
  \centering
  \includegraphics[height=10cm,width=16cm]{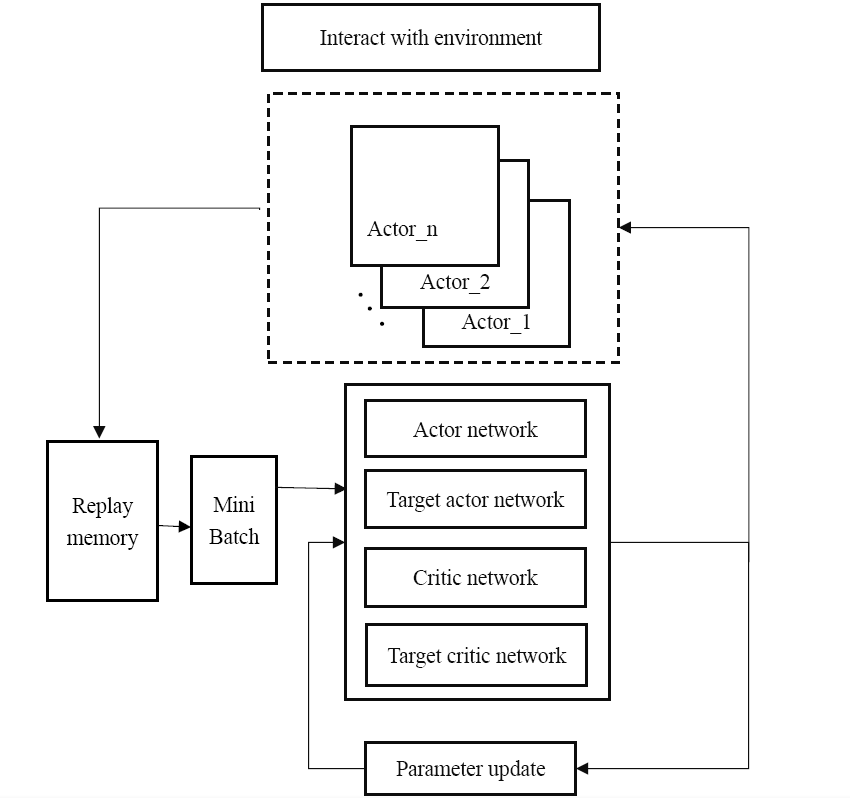}
  \caption{The general architecture of the proposed model}
  \label{architecture}
\end{figure*}

We use a multiprocessing method to implement  Ape-X. Specifically, there are 22 parallel processes, of which 20 are responsible for interacting with the environment, one process is accountable for updating network parameters, and one process is responsible for data exchange.
The general architecture of the proposed method and the pseudocode for the algorithm are shown in Fig.\ref{architecture} and Table \ref{Pseudocode}.\par
\begin{table*}[t]
\centering
\caption{Pseudocode of the proposed method} \label{Pseudocode}
\setlength{\tabcolsep}{4mm}{
\begin{tabular}{l}
  \hline
  $\mathbf{Algorithm 1. \enspace R3L}$\\
  \hline
  $\mathbf{Input:}$ bath size M, number of actors K, replay size R, exploration constant, learning rate   \\
  1: Initialize network weights $(\varphi,\omega)$ \\
  2: Launch K actors and replicate network weights $\varphi$ to each actor\\
  4:$\enspace$$\mathbf{for}$ t=1,2,...,T, $\mathbf{do}$ \\
  5:\quad \quad Sample M transitions $(s_k,a_k,s'_k)$ of length N from replay buffer\\
  6:\quad \quad Compute the N quantiles of $Z(s,a)$: $[\theta_{1,k}, \theta_ {2,k}, \ldots, \theta_ {N,k}]=K_{\omega}(s_{k},a_{k})$\\
  6:\quad \quad Construct the target distributions $[\tau \theta'_{1,k},\tau \theta'_{2,k}, ...\tau \theta'_{N,k}]=r+\gamma K_{\omega}(s'_k,\mu(s'_k))$\\
  7:\quad \quad Update the parameters of critic network($\omega$) by minimize: $loss=\frac{1}{M}\sum_{k}L^{k}_{\omega}$\\
  8:\quad \quad Update the parameters of critic network($\varphi$) by gradient ascend:\\
    \quad \quad \quad \quad \quad \quad$\nabla _{\varphi} \approx \frac{1}{M}\sum_{k}\nabla_{a_k}U(s_k,a_k) \nabla_{\varphi} \mu_{}(s_{k})$\\
  9:\quad \quad If $t=0$ mod $t_{target}$, update the target network:\\
  \quad \quad \quad \quad \quad \quad $\varphi\leftarrow \tau \varphi+(1-\tau)\varphi'$, $\omega\leftarrow \tau \omega+(1-\tau)\omega'$\\

  10: \quad \enspace If $t=0$ mod $t_{actor}$, replicate network weights to the actors\\
  11:$\enspace$$\mathbf{end \enspace for}$\\
  12:$\enspace$$\mathbf{return}$ actor network parameters $\varphi$\\
  $\mathbf{Actor}$ \\
  \hline
  1:$\enspace$$\mathbf{repeat}$\\
  2:\quad Sample action $a=\mu_{\varphi}(s)+\epsilon$\\
  3:\quad Execute action $a$, observe reward r and state $s'$\\
  4:\quad Store $(s,a,r,s')$ in replay buffer\\
  5:$\enspace$$\mathbf{until}$ learner finishes\\

  \hline
\end{tabular}}
\end{table*}\par
\section[5]{Experimental setup}\label{section5}
This section details the setups of our experiments, including datasets, performance measures, benchmark models, and technical details.
In addition, we also need to make the following assumptions:
Firstly, all transactions are made at the close price at the end of each trading period;  secondly, the market size is large enough that the price of security and market environment is not affected by the transactions;
Thirdly, since frequent adjustment of portfolio weights will generate a lot of transaction costs, we adjust the portfolio weights once a week; finally, the investment period is set to one year.
\subsection[5.1]{Datasets}
We experiment with two different portfolios. The first portfolio consists of four famous exchange trading funds (ETFs) in the US market and a risk-free asset. The ETFs portfolio includes SPDR S\&P 500 ETF Trust (SPY), Invesco QQQ Trust ETF(QQQ), SPDR Dow Jones Industrial Average ETF(DIA), and iShares Russell 2000 ETF(IWM). The second portfolio consists of stocks of four technology companies and a risk-free asset. The four technology companies are ORCL, AAPL, TSLA, and GOOG. All data used in this paper is available on Yahoo Finance. The trading horizon of 13 years is divided into both training and testing sets as following:\par
$\bullet$ $\mathbf{Training\enspace set:}$ $01/01/2008\rightarrow 31/12/2018.$\par
$\bullet$ $\mathbf{testing \enspace set:}$  $01/01/2019\rightarrow 31/12/2022.$\par
\subsection[5.2]{Performance measures}
Risk and return are inseparable in the investment decision process. Under normal circumstances, the risk is highly correlated with the return, and a high return means high risk. Therefore, the performance measures must include these two aspects. In this article, we use four types of performance measures to evaluate the proposed trading strategy. The first type of metric measures the profitability of the investment strategy, i.e., total return. The second type of metrics measures investment risk, including variance and VaR. The third metric type considers risk and returns, including sharpe ratio and Sortino Ratio. The last type of metric is average turnover. More detail about performance measures are as following.\par
$\bullet$ $\mathbf{Total \enspace return(TR).}$ The total return is the rate of return over a period. Total return, computed using Eq.(\ref{equation:2}), includes capital gains, interest, realized distributions, and dividends.
\begin{equation}\label{equation:2}
 TR=(Q_{[T]}-Q_{[0]})/Q_{[0]}
\end{equation}
Where $Q_{[0]}$ is the value of the initial investment;$Q_{[T]}$ is the value of the portfolio at the end of the investment period.\par
$\bullet$ $\mathbf{Value \enspace at \enspace risk(VaR).}$ As mentioned above, VaR calculates the maximum loss of a portfolio over a given period on specified confidence level.\par
$\bullet$ $\mathbf{Sharpe \enspace ratio(SR1).}$ The Sharpe ratio, first proposed by cite{sharpe1998sharpe}, is a measure of risk-adjusted return. This ratio, computed using Eq.(\ref{sharp}), reflects how much the return on risk per unit exceeds the risk-free return. If the Sharpe ratio is positive, a portfolio's average net value growth rate exceeds the risk-free interest rate.
\begin{equation}\label{sharp}
 SR=(E[R_{p}]-R_{f})/\sigma_{p}
\end{equation}
where $E[R_{p}]$ is the expectation of portfolio return, $R_{f}$ is  risk-free rate, $\sigma_{p}$ is standard deviation of portfolio return.\par
 $\bullet$ $\mathbf{Sortino \enspace ratio(SR2).}$ The Sortino ratio is a risk-adjustment metric used to determine the additional return for each unit of downside risk. It is similar to the Sharpe ratio, but the Sortino ratio uses the lower partial standard deviation rather than the total standard deviation to distinguish between adverse and favorable fluctuations. The Sortino ratio can be expressed as follows:
 \begin{equation}\label{equation:12}
 SR=(E[R_{p}]-R_{f})/DR
\end{equation}
where $DR$ is the lower partial standard deviation of portfolio return.\par
$\bullet$ $\mathbf{Standard \enspace deviation(SD).}$ The standard deviation of the portfolio variance can be calculated as the square root of the portfolio variance.\par
$\bullet$ $\mathbf{Average \enspace turnover(AT).}$ Average turnover represents the average level of change in portfolio weight, which is defined in Eq.(\ref{equation:13}).
\begin{equation}\label{equation:13}
 AT=1/(2t_{f})\sum_{t=0}^{t_{f}-1}\sum_{i=1}^{l}\mid{w_{t+1,i}-w_{t,i}}\mid
\end{equation}
where $t_{f}$ is the investment horizon,$w_{t,l}$ denote weight parameter of asset i in investment period t.

\subsection[5.3]{Benchmark models}
To analyze the effectiveness of the proposed strategy, some benchmark strategies, summarised hereafter, are selected for comparison.\par
$\bullet$ $\mathbf{Buy \enspace and \enspace hold(B\&H).}$ B\&H is used as a benchmark strategy by many researchers compared with their proposed strategies. Suppose that the holding proportion of all five assets is 20\% in the B\&H strategy and remains unchanged throughout the investment period.\par
$\bullet$ $\mathbf{Sell \enspace and \enspace hold(S\&H).}$ S\&H is also widely used as a benchmark strategy. Assuming that in the S\&H strategy, the holding proportion of all four risky assets is -25\%, the proportion of risky-free assets is 200\%, all of which remain unchanged throughout the investment period.\par
$\bullet$ $\mathbf{Random \enspace selected (RN).}$  According to the efficient market hypothesis (EMH), all valuable information has been timely, accurate, and fully reflected in the stock price trend, including the current and future value of the enterprise. Without market manipulation, investors cannot obtain excess profits higher than the market average by analyzing past prices. Any trading strategy based on historical data differs from the randomly selected strategy. \par
$\bullet$ $\mathbf{Mean-variance \enspace model.}$ The mean-variance model, introduced by Markowitz in 1952, aims to find the best portfolio only by the first two moments of cumulative return. Suppose there are n kinds of assets, $R=(r_1,...r_n)^T$ represents the expected return of a portfolio,  $W=(W_1,...W_n)^T$  is the weight vector, $\Sigma$ is the covariance vector of return, $\zeta$ is risk aversion coefficient, $\mathbf{1}$ represents $n\times1 $ dimensional unit vector, we establish the following optimization model based on utility maximization:
\begin{eqnarray}
  max \enspace U= &W^{T}R-\zeta W^{T}\Sigma W\\
s.t.& 1^T W = 1
\end{eqnarray}\par
\subsection[5.4]{technique detail}
We obtain time window size and other hyper-parameters, including replay buffer size, batch size, discount factor, etc., through several rounds of tuning. In addition, we need to set some other hyper-parameters before training. Suppose the risk-free rate of return is 2\%, and the equivalent weekly return is 0.038\%. The transaction cost of buying and selling an asset is set to 0.02\%. Risk aversion parameter($\zeta$) and short selling parameter($\delta$)are set to 0.5 and 3; because these two parameters have significant impacts on investment decisions and portfolio returns, we perform sensitivity analysis for them in subsection \ref{section6.3}. The networks were trained through the ADAM optimization procedure with a learning rate of $10^{-5}$. The activation function is set as the Leaky Relu function. All parameter are summarised in Table \ref{hyper-parameters}. Finally, the algorithms proposed in the paper are implemented in python3.7.11  using pytorch1.10.2 and were run on a PC that contains a sixteen-core 2.50GHz CPU with  16GB RAM and NVIDIA Geforce RTX 3060 GPU.\par
\begin{table*}[t]
\centering
\caption{summary of hyper-parameters} \label{hyper-parameters}
\setlength{\tabcolsep}{4mm}{
\begin{tabular}{l l| l l}
  \hline
  hyper-parameter    & value        & hyper-parameter           & value \\
  \hline
  time window size(n)    & 60        & replay memory            & 2000 \\

  learning rate(lr)      & 1e-5      & number of parameter update        & 80000 \\

  batch size             & 32        & discount factor($\gamma$)          &  0.9 \\

  confidence level of VaR($\alpha$)             & 0.95  & parameter of Huber Loss($\kappa$)  & 1.0 \\

  short selling parameter($\delta$)    & 3.0   & soft update parameter($\tau$)    & 0.5 \\
  initial money           &10000.   & risk-free rate     &3.8e-4 \\
  risk attitude parameter($\zeta$)           & 0.5     & num of layer for GRU  &2\\
  Output number of critic network &200 & transaction cost &0.0020 \\

  \hline
\end{tabular}}
\end{table*}\par
\section[6]{Result and discussion}\label{section6}
\subsection[6.1]{The ETFs portfolio}
\begin{figure*}[t]
  \centering
  \includegraphics[height=10cm,width=18cm]{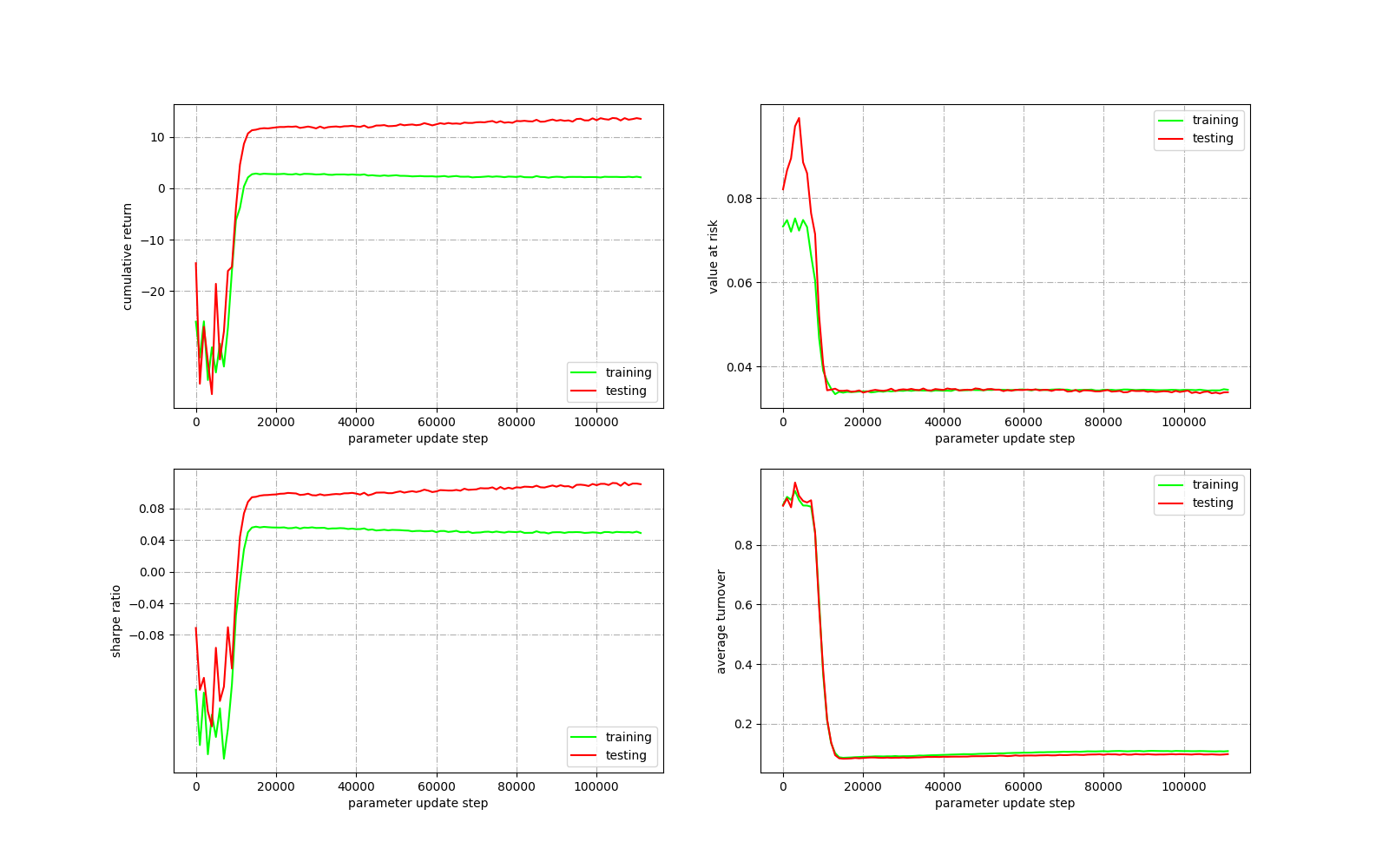}
  \caption{Average performance for the ETFs portfolio}
  \label{ETF_TT}
\end{figure*}

The first detailed analysis concerns the execution of the proposed algorithm on the ETFs portfolio. Fig.\ref{ETF_TT} illustrates the average performance of the proposed algorithms for both the training and testing sets as the parameter update step continues. It can be noticed that the cumulative return of the ETFs portfolio tends to converge in the training and testing set after about 20000 parameter updates. Moreover, we note that the convergence value of the cumulative return in the testing set is greater than in the training set. On the other hand, the risk levels measured by VaR are almost identical in training and testing sets. One possible explanation is that, as pointed out by\cite{theate2021application}, because the training and testing set does not share identical distributions, it simply indicates an easier-to-trade and more profitable market for the testing set. Although the training and testing sets do not share identical distributions, the training set is representative enough for the R3L algorithm to achieve good results in testing, so we can still use historical data to train the model and then use it for future trading. Finally, from the perspective of the Sharpe ratio, the performance of the R3L algorithm in the testing set is still better than that in the training set.\par
\begin{figure*}[t]
\centering
  \setlength{\abovecaptionskip}{0.cm}
  \setlength{\belowcaptionskip}{-0cm}
\includegraphics[height=6cm,width=18cm]{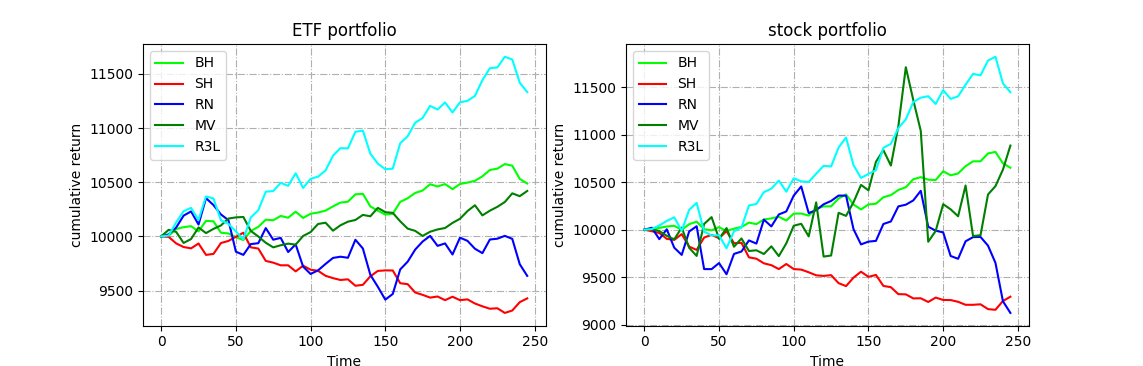}
\caption{Performance of R3L algorithm}
\label{ETF}
\end{figure*}

Fig.\ref{ETF} illustrates the portfolio value trend when applying the proposed algorithm and benchmark strategies in the testing set. We observe that the final portfolio value of the proposed algorithm is 19.51\%  higher than the S\&H strategy, 18.0\% higher than the RN strategy, 8.1\% higher than the MV strategy, and 7.9\% higher than the B\&H strategy.
\par
Table \ref{ETF_TABLE} further presents output performance measures results when using the R3L algorithm and the benchmark strategies. The R3L algorithm is optimal regarding risk, return, and overall performance. The Sharpe ratio of the proposed algorithm is 19.51\%  higher than the S\&H strategy, 18.0\% higher than the RN strategy, 8.1\% higher than the MV strategy, and 18.0\% higher than the B\&H strategy.
In addition, as the stock market was primarily bullish throughout the test period, the B\&H strategy outperformed other benchmark strategies most of the time,  while the performance of the S\&H strategy was the worst.
The performance of the RN strategy is abysmal. Since we did not consider the company's financial and internal non-public information, it does not mean that the efficient market hypothesis is not tenable.
\begin{table*}[t]
\centering
  \setlength{\abovecaptionskip}{0.cm}
  \setlength{\belowcaptionskip}{-0cm}
\caption{Main result} \label{ETF_TABLE}
\setlength{\tabcolsep}{4mm}{
\begin{tabular}{l|l|l|l|l|l|l|l}
\hline

  & delta &TR  &SD &SR1 &VAR &SR2 &AT\\
  \hline
  &B\&H &5.00\%	&0.0264 	&0.0889 	&0.0374 	&0.1260 	&0.0000 	\\
  &S\&H &-5.24\%	&0.0330 	&-0.0870 	&0.0419 	&-0.1392 	&0.0000 	\\
  &RN &-4.00\%	&0.0329 	&-0.0414 	&0.0435 	&-0.0541 	&0.3291 	\\
  &MV &4.86\%	&0.0605 	&0.0968 	&0.0875 	&0.1128 	&0.0929 	\\
  &R3L &13.32\%	&0.0266 	&0.1100 	&0.0338 	&0.1681 	&0.0966 	\\
\hline
  &B\&H &10.09\%	&0.0317 	&0.0780 	&0.0203 	&0.1173 	&0.0000 	\\
  &S\&H &-20.62\%	&0.0396 	&-0.1280 	&0.0575 	&-0.1657 	&0.0000 	\\
  &RN &-8.76\%	&0.0423 	&-0.0469 	&0.0584 	&-0.0536 	&0.3046 	\\
  &MV &6.75\%	&0.0947 	&0.0893 	&0.0115 	&0.1718 	&0.0929 	\\
  &R3L &14.47\%	&0.0298 	&0.0943 	&0.0369 	&0.1571 	&0.0849 	\\
\hline
\end{tabular}}
\end{table*}\par
\subsection[6.2]{The stock portfolio}
\begin{figure*}[t]
\centering
\includegraphics[height=10cm,width=18cm]{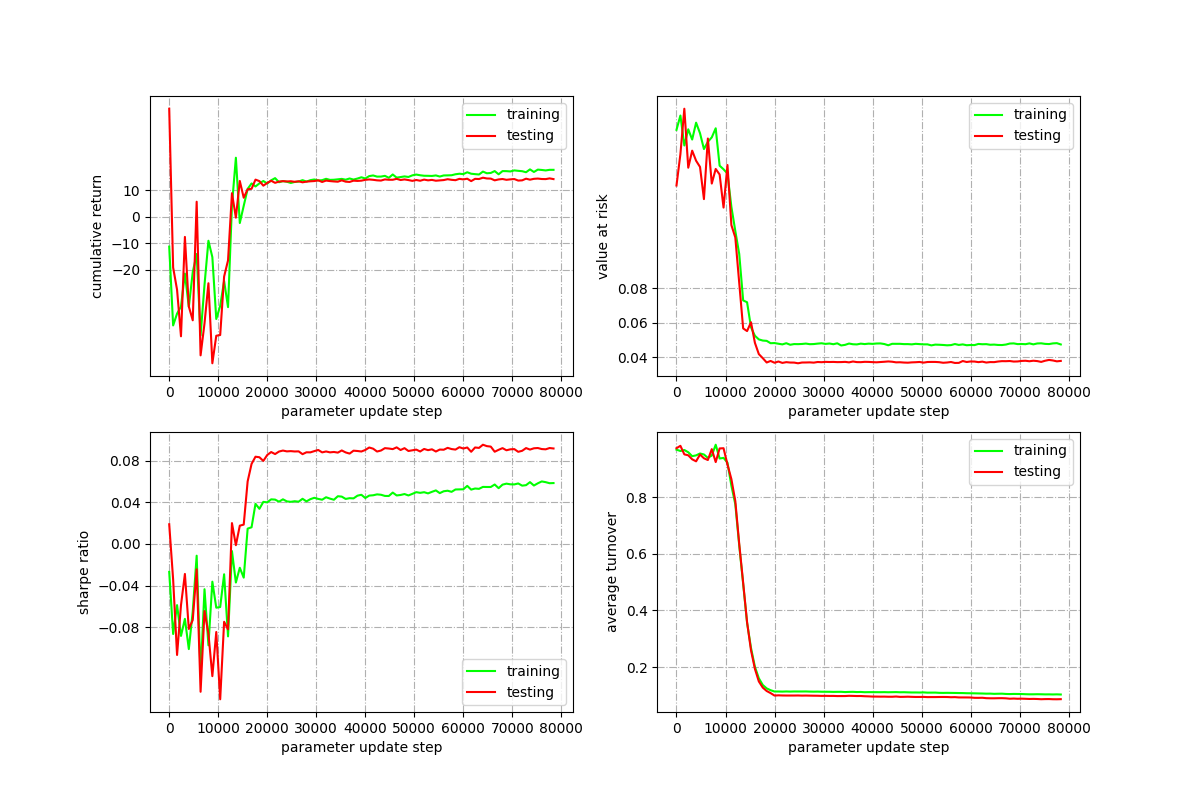}
\caption{Average performance  for the stock portfolio}
\label{stock}
\end{figure*}

The same detailed analysis is performed on the stock portfolio, which shows different characteristics compared to the ETFs portfolio. Fig.\ref{stock} illustrates the average performance of the proposed algorithm for both the training and testing sets. We observe that the cumulative return of the stock portfolio is almost identical in the training and testing set after 20000 parameter updates. At the same time, the risk level(VaR) is higher in training sets than in testing sets, which is very different from the ETFs portfolio. Finally, from the sharpe ratio perspective, the algorithm's performance in the training set is better than in the testing set.
Fig.\ref{ETF} illustrates the portfolio value trend of different strategies. Similar to the ETFs portfolio, the R3L strategy has the highest cumulative value at the end of the investment period.
Table \ref{ETF_TABLE} presents output performance measures results when using different trading strategies. It can be noticed that from the perspective of the sharpe ratio, sortino ratio, and VaR, the R3L algorithm is also optimal.
\subsection[6.3]{Sensitive analysis}\label{section6.3}
In this paper, delta($\delta$) and zeta($\zeta$) are two critical parameters that affect the overall performance of the proposed algorithm. Delta represents the maximum ratio of short selling allowed; the larger the delta, the more profit opportunities investors have in the bear market. Zeta represents investors' risk attitude; the larger zeta is, the higher the risk tolerance is, and investors tend to adopt more radical investment strategies. Finally, as mentioned in subsection \ref{section4.2}, different network structures have different feature extraction abilities. Therefore, it is necessary to analyze the impact of network structures on the algorithm's effectiveness to select the best one. We conduct the sensitive analysis of delta, theta, and network structure on the R3L strategy. The performance of the proposed model is evaluated concerning different values for delta, theta, and network structures.\par

\begin{figure*}[t]
\centering
  \setlength{\abovecaptionskip}{0.cm}
  \setlength{\belowcaptionskip}{-0cm}
\includegraphics[height=6cm,width=18cm]{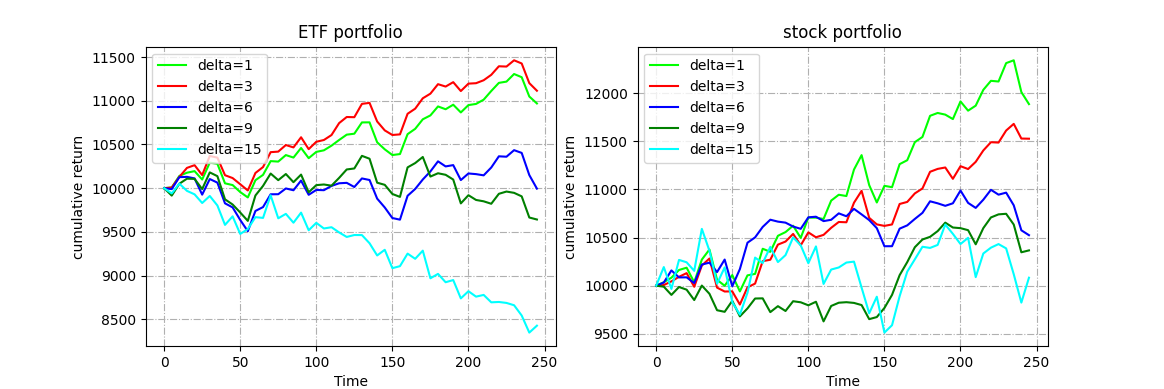}
\caption{Impact of delta value on R3L}
\label{delta}
\end{figure*}

$\bullet$ $\mathbf{Sensitive \enspace analysis\enspace of\enspace delta.}$ Fig.\ref{delta} illustrates the portfolio value trend when applying the R3L algorithm with a different delta. Portfolio value shows an upward trend over time for the ETFs and stock portfolios when delta equals 1 and 3. In this case, investors can obtain positive returns at the end of the investment period. In contrast, in other cases, the portfolio value shows significant volatility over time, and investors receive negative returns at the end of the investment period. Table \ref{delta_table} further shows the overall performance of the proposed strategy concerning different delta values. From the perspective of total return, Sharpe ratio, and Sortino ratio, $\delta=3$ is optimal for the ETFs portfolio, and $\delta=1$ is optimal for the stock portfolio.\par
It can be seen that the maximum amount of short selling allowed is not the bigger, the better, which is explained by the fact that, although short selling can provide investors with profit opportunities in a bear market, it also brings more risks. Therefore, it is essential to control the scale of short selling appropriately.
\par
\begin{table*}[t]
\centering
  \setlength{\abovecaptionskip}{0.cm}
  \setlength{\belowcaptionskip}{-0cm}
\caption{Sensitive analyse of delta} \label{delta_table}
\setlength{\tabcolsep}{4mm}{
\begin{tabular}{l|l|l|l|l|l|l|l}
\hline
  & delta &TR  &SD &SR1 &VAR &SR2 &AT \\
  \hline
  &1 &9.70\%	&0.0266 	&0.0815 	&0.0367 	&0.1177 	&0.0247 	\\
  &3 &11.16\%	&0.0270 	&0.0912 	&0.0362 	&0.1391 	&0.1029 	\\
  &6 &-0.05\%	&0.0323 	&0.0060 	&0.0484 	&0.0183 	&0.2362 	\\
  &9 &-3.59\%	&0.0349 	&-0.0346 	&0.0605 	&-0.0452 	&0.3681 	\\
  &15 &-15.74\%	&0.0383 	&-0.1034 	&0.0620 	&-0.1302 	&0.6059 	\\
\hline

  &1 &18.85\%	&0.0332 	&0.1117 	&0.0432 	&0.1962 	&0.0288 	\\
  &3 &15.26\%	&0.0288 	&0.0991 	&0.0346 	&0.1834 	&0.0941 	\\
  &6 &5.24\%	&0.0276 	&0.0220 	&0.0344 	&0.0827 	&0.1995 	\\
  &9 &3.66\%	&0.0350 	&0.0300 	&0.0478 	&0.0514 	&0.2721 	\\
  &15 &0.83\%	&0.0532 	&-0.0095 	&0.0845 	&-0.0170 	&0.4229 	\\
\hline
\end{tabular}}
\end{table*}\par
\begin{figure*}[t]
\centering
  \setlength{\abovecaptionskip}{0.cm}
  \setlength{\belowcaptionskip}{0.cm}
\includegraphics[height=6cm,width=18cm]{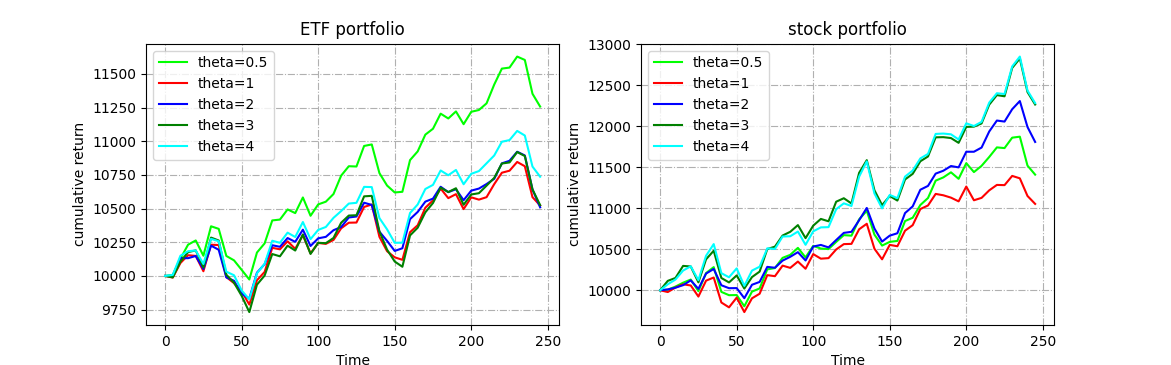}
\caption{Impact of theta value on R3L}
\label{zeta}
\end{figure*}

$\bullet$ $\mathbf{Sensitive \enspace analysis\enspace of\enspace zeta.}$  Fig.\ref{zeta} illustrates the portfolio value trend when applying the R3L algorithm with different zeta. Table \ref{zeta_table} further shows the overall performance of the proposed strategy. It can be noticed that the change in portfolio value over time is similar for different theta. At the end of the investment period, the portfolio value is the greatest for the ETFs portfolio when theta equals 3. From the Sharpe and Sortino ratios perspective, $\zeta=3$ is also optimal. The portfolio value is the greatest for the stock portfolio when theta equals 4. From the standpoint of Sharpe ratio and Sortino ratio, $\zeta=4$ is also optimal. Nevertheless, the VaR is the lowest when theta equals 2 for the ETFs and stock portfolios.\par
From the above analysis, it can be seen that a higher risk aversion coefficient does not necessarily lead to a smaller return and risk level, which contradicts the classical portfolio theory. One possible explanation is that our algorithm can predict the stock price based on historical data to improve the portfolio's return and reduce the risk. To sum up, the algorithm proposed in this article can enhance portfolio returns and minimize risk.

\par
\begin{table*}[t]
\centering
  \setlength{\abovecaptionskip}{0.cm}
  \setlength{\belowcaptionskip}{0.cm}
\caption{Sensitive analyse of zeta} \label{zeta_table}
\setlength{\tabcolsep}{4mm}{
\begin{tabular}{l|l|l|l|l|l|l|l}
\hline

  & zeta &TR  &SD &SR1 &VAR &SR2 &AT \\
  \hline
  &0.5 &12.58\%	&0.0267 	&0.1029 	&0.0341 	&0.1576 	&0.0991 	\\
  &1.0 &5.23\%	&0.0275 	&0.0453 	&0.0383 	&0.0703 	&0.1122 	\\
  &2.0 &5.10\%	&0.0270 	&0.0486 	&0.0395 	&0.0814 	&0.1141 	\\
  &3.0 &5.26\%	&0.0305 	&0.0521 	&0.0432 	&0.0854 	&0.1019 	\\
  &4.0 &7.37\%	&0.0264 	&0.0570 	&0.0401 	&0.0887 	&0.0955 	\\
\hline

  &0.5 &14.11\%	&0.0297 	&0.0898 	&0.0372 	&0.1548 	&0.0979 	\\
  &1.0 &10.53\%	&0.0281 	&0.0649 	&0.0383 	&0.1169 	&0.1156 	\\
  &2.0 &18.09\%	&0.0253 	&0.1295 	&0.0344 	&0.2470 	&0.0788 	\\
  &3.0 &22.66\%	&0.0338 	&0.1236 	&0.0448 	&0.2623 	&0.0867 	\\
  &4.0 &22.83\%	&0.0351 	&0.1247 	&0.0445 	&0.2450 	&0.0835 	\\
\hline
\end{tabular}}
\end{table*}\par
\begin{figure*}[t]
\centering
  \setlength{\abovecaptionskip}{0.cm}
  \setlength{\belowcaptionskip}{0.cm}
\includegraphics[height=6cm,width=18cm]{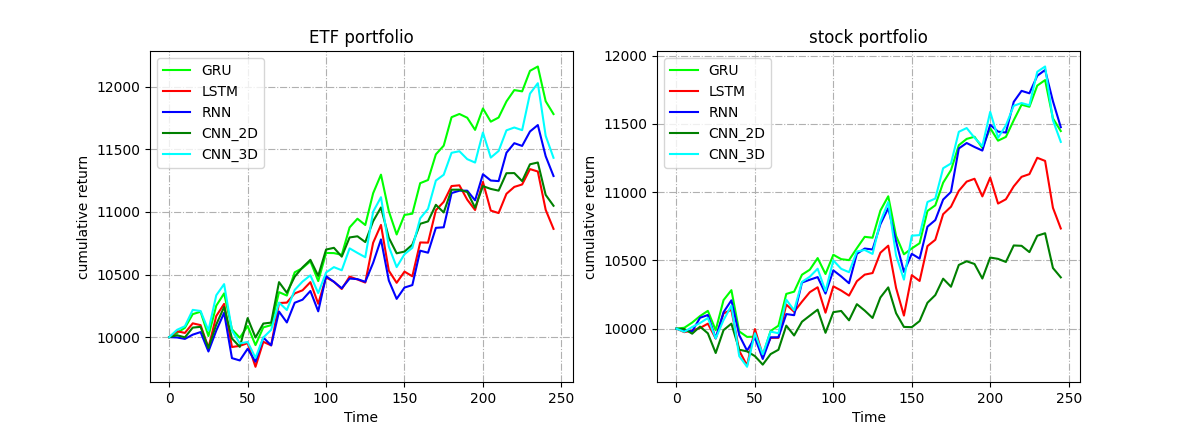}
\caption{Impact of network structure on R3L}
\label{network_fig}
\end{figure*}

$\bullet$ $\mathbf{Sensitive \enspace analysis\enspace of\enspace network \enspace structure.}$ Fig. \ref{network_fig} illustrates the portfolio value trend when applying the R3L algorithm with a different network structure. It can be observed the changing trend of portfolio value is consistent under different network structures, and only the final cumulative value is slightly different. GRU and RNN obtained the maximum cumulative value for the ETFs and stock portfolios. Table \ref{network_table} further shows the overall performance. We find that  GRU performs best in the ETFs portfolio from the perspective of Sharpe ratio and Sortino ratio, and RNN  performs best in the stock portfolio from the perspective of Sortino ratio. In contrast, GRU performs best from the perspective of the sharpe ratio.
\par
\begin{table*}[t]
\centering
  \setlength{\abovecaptionskip}{0.cm}
  \setlength{\belowcaptionskip}{0.cm}
\caption{Sensitive analyse of network} \label{network_table}
\setlength{\tabcolsep}{4mm}{
\begin{tabular}{l|l|l|l|l|l|l|l}
\hline

  & Network &TR  &SD &SR1 &VAR &SR2 &AT \\
  \hline
  &GRU &17.82\%	&0.0342 	&0.1089 	&0.0457 	&0.2118 	&0.0865 	\\
  &LSTM &8.65\%	&0.0337 	&0.0433 	&0.0531 	&0.0723 	&0.1060 	\\
  &RNN &12.88\%	&0.0335 	&0.0660 	&0.0455 	&0.1173 	&0.1093 	\\
  &COONV2D &10.50\%	&0.0321 	&0.0680 	&0.0355 	&0.1381 	&0.1258 	\\
  &CONV3D &14.33\%	&0.0352 	&0.0758 	&0.0526 	&0.1584 	&0.0985 	\\
\hline
 
  &GRU &14.47\%	&0.0298 	&0.0943 	&0.0369 	&0.1571 	&0.0849 	\\
  &LSTM &7.34\%	&0.0357 	&0.0594 	&0.0534 	&0.0917 	&0.1110 	\\
  &RNN &14.75\%	&0.0300 	&0.0907 	&0.0431 	&0.2078 	&0.1125 	\\
  &CONV2D &3.76\%	&0.0283 	&0.0377 	&0.0326 	&0.0840 	&0.1541 	\\
  &CONV3D &13.67\%	&0.0360 	&0.0710 	&0.0521 	&0.1460 	&0.0903 	\\
\hline
\end{tabular}}
\end{table*}\par
\section[7]{Conclusion and future work}\label{section7}
We propose a novel deep reinforcement learning algorithm, called risk-return reinforcement learning(R3L), to derive a portfolio trading strategy. Compared with the existing literature, the main innovations of this paper are as the following: firstly, we construct a portfolio optimization model which is solved by an improved deep reinforcement learning algorithm based on actor-critic architecture; secondly, we realize short selling of portfolio through a linear transformation; thirdly, we leverage Ape-x algorithm to speed up the training process. Experiments carried out on the performance of the R3L algorithm demonstrate that the proposed R3L is superior to the traditional benchmark strategies, such as buy-and-hold, sell-and-hold, random select, and mean-variance strategies. In addition, although short selling allows investors to profit in a bear market, the maximum short selling permitted ratio is not the bigger, the better. Therefore, investors should choose optimal short selling parameters according to investment objectives, asset types, and other factors to maximize portfolio return. Similarly, we need to choose appropriate risk attitude parameters and network structure to optimize the portfolio's overall performance. \par
Future research can be carried out from the following aspects:
First, according to to\cite{rockafellar2000optimization}, VaR is not a coherent risk measure, so we could consider using other risk measurements, such as conditional value at risk(CVaR), to construct the portfolio optimization problem in future research;
Second, this paper assumes that the trading volume is small compared to the market size, so the trading behavior of a single agent does not affect the market environment and stock prices. However, the trading behavior, even with a small volume,  still has a subtle impact on the stock price and market environment. How to model the market environment is one of the future research directions;
Finally, the decision-making process of algorithmic trading can be divided into two parts: selecting asset types and determining optimal portfolio weight. Therefore, subsequent research can be used hierarchical deep reinforcement learning(HDRL) to handle portfolio optimization problems.\par
\section*{Conflict of interest}
The authors declare that they have no conflict of interests regarding the publication of this paper.

\section*{Acknowledgments}
Here are the acknowledgments. Note the asterisk \textbackslash{\ttfamily section*}\{{\ttfamily Acknowledgments}\} that signifies that this section is unnumbered.

\bibliographystyle{pisikabst}
\bibliography{bibfile}

\begin{thebibliography}{42}
\expandafter\ifx\csname natexlab\endcsname\relax\def\natexlab#1{#1}\fi
\expandafter\ifx\csname bibnamefont\endcsname\relax
  \def\bibnamefont#1{#1}\fi
\expandafter\ifx\csname bibfnamefont\endcsname\relax
  \def\bibfnamefont#1{#1}\fi
\expandafter\ifx\csname citenamefont\endcsname\relax
  \def\citenamefont#1{#1}\fi
\expandafter\ifx\csname url\endcsname\relax
  \def\url#1{\texttt{#1}}\fi
\expandafter\ifx\csname urlprefix\endcsname\relax\def\urlprefix{}\fi
\providecommand{\bibinfo}[2]{#2}
\providecommand{\eprint}[1]{\href{http://arxiv.org/abs/#1}{arXiv:#1}}

\bibitem[{\citenamefont{Perold}(1988)}]{perold1988implementation}
\bibinfo{author}{\bibfnamefont{A.~F.} \bibnamefont{Perold}},
  \bibinfo{title}{The implementation shortfall: Paper versus reality},
  \emph{\bibinfo{journal}{Journal of Portfolio Management}}
  \textbf{\bibinfo{volume}{14}}, \bibinfo{pages}{4} (\bibinfo{year}{1988}).

\bibitem[{\citenamefont{Almgren and Lorenz}(2007)}]{almgren2007adaptive}
\bibinfo{author}{\bibfnamefont{R.}~\bibnamefont{Almgren}} \bibnamefont{and}
  \bibinfo{author}{\bibfnamefont{J.}~\bibnamefont{Lorenz}},
  \bibinfo{title}{Adaptive arrival price}, \emph{\bibinfo{journal}{Trading}}
  \textbf{\bibinfo{volume}{2007}}, \bibinfo{pages}{59} (\bibinfo{year}{2007}).

\bibitem[{\citenamefont{Berkowitz et~al.}(1988)\citenamefont{Berkowitz, Logue,
  and Noser~Jr}}]{berkowitz1988total}
\bibinfo{author}{\bibfnamefont{S.~A.} \bibnamefont{Berkowitz}},
  \bibinfo{author}{\bibfnamefont{D.~E.} \bibnamefont{Logue}}, \bibnamefont{and}
  \bibinfo{author}{\bibfnamefont{E.~A.} \bibnamefont{Noser~Jr}},
  \bibinfo{title}{The total cost of transactions on the nyse},
  \emph{\bibinfo{journal}{The Journal of Finance}}
  \textbf{\bibinfo{volume}{43}}, \bibinfo{pages}{97} (\bibinfo{year}{1988}).

\bibitem[{\citenamefont{Madhavan and Panchapagesan}(2002)}]{madhavan2002first}
\bibinfo{author}{\bibfnamefont{A.~N.} \bibnamefont{Madhavan}} \bibnamefont{and}
  \bibinfo{author}{\bibfnamefont{V.}~\bibnamefont{Panchapagesan}},
  \bibinfo{title}{The first price of the day}, \emph{\bibinfo{journal}{The
  Journal of Portfolio Management}} \textbf{\bibinfo{volume}{28}},
  \bibinfo{pages}{101} (\bibinfo{year}{2002}).

\bibitem[{\citenamefont{Kolm and Maclin}(2011)}]{kolm2011algorithmic}
\bibinfo{author}{\bibfnamefont{P.~N.} \bibnamefont{Kolm}} \bibnamefont{and}
  \bibinfo{author}{\bibfnamefont{L.}~\bibnamefont{Maclin}},
  \bibinfo{title}{Algorithmic trading, optimal execution, and dyna mic port
  folios} (\bibinfo{year}{2011}).

\bibitem[{\citenamefont{Kritzman}(2006)}]{kritzman2006optimizers}
\bibinfo{author}{\bibfnamefont{M.}~\bibnamefont{Kritzman}}, \bibinfo{title}{Are
  optimizers error maximizers?}, \emph{\bibinfo{journal}{The Journal of
  Portfolio Management}} \textbf{\bibinfo{volume}{32}}, \bibinfo{pages}{66}
  (\bibinfo{year}{2006}).

\bibitem[{\citenamefont{Hendershott and
  Riordan}(2013)}]{hendershott2013algorithmic}
\bibinfo{author}{\bibfnamefont{T.}~\bibnamefont{Hendershott}} \bibnamefont{and}
  \bibinfo{author}{\bibfnamefont{R.}~\bibnamefont{Riordan}},
  \bibinfo{title}{Algorithmic trading and the market for liquidity},
  \emph{\bibinfo{journal}{Journal of Financial and Quantitative Analysis}}
  \textbf{\bibinfo{volume}{48}}, \bibinfo{pages}{1001} (\bibinfo{year}{2013}).

\bibitem[{\citenamefont{Eriksson and Swartling}(2012)}]{eriksson2012ucd}
\bibinfo{author}{\bibfnamefont{E.}~\bibnamefont{Eriksson}} \bibnamefont{and}
  \bibinfo{author}{\bibfnamefont{A.}~\bibnamefont{Swartling}}, in
  \emph{\bibinfo{booktitle}{Human Work Interaction Design-HWID2012}}
  (\bibinfo{year}{2012}),  \bibinfo{pages}{116--126}.

\bibitem[{\citenamefont{Jondeau and Rockinger}(2003)}]{jondeau2003testing}
\bibinfo{author}{\bibfnamefont{E.}~\bibnamefont{Jondeau}} \bibnamefont{and}
  \bibinfo{author}{\bibfnamefont{M.}~\bibnamefont{Rockinger}},
  \bibinfo{title}{Testing for differences in the tails of stock-market
  returns}, \emph{\bibinfo{journal}{Journal of Empirical Finance}}
  \textbf{\bibinfo{volume}{10}}, \bibinfo{pages}{559} (\bibinfo{year}{2003}).

\bibitem[{\citenamefont{Tsiakas}(2006)}]{tsiakas2006periodic}
\bibinfo{author}{\bibfnamefont{I.}~\bibnamefont{Tsiakas}},
  \bibinfo{title}{Periodic stochastic volatility and fat tails},
  \emph{\bibinfo{journal}{Journal of Financial Econometrics}}
  \textbf{\bibinfo{volume}{4}}, \bibinfo{pages}{90} (\bibinfo{year}{2006}).

\bibitem[{\citenamefont{Franke}(2009)}]{franke2009applying}
\bibinfo{author}{\bibfnamefont{R.}~\bibnamefont{Franke}},
  \bibinfo{title}{Applying the method of simulated moments to estimate a small
  agent-based asset pricing model}, \emph{\bibinfo{journal}{Journal of
  Empirical Finance}} \textbf{\bibinfo{volume}{16}}, \bibinfo{pages}{804}
  (\bibinfo{year}{2009}).

\bibitem[{\citenamefont{Park et~al.}(2020)\citenamefont{Park, Sim, and
  Choi}}]{park2020intelligent}
\bibinfo{author}{\bibfnamefont{H.}~\bibnamefont{Park}},
  \bibinfo{author}{\bibfnamefont{M.~K.} \bibnamefont{Sim}}, \bibnamefont{and}
  \bibinfo{author}{\bibfnamefont{D.~G.} \bibnamefont{Choi}}, \bibinfo{title}{An
  intelligent financial portfolio trading strategy using deep q-learning},
  \emph{\bibinfo{journal}{Expert Systems with Applications}}
  \textbf{\bibinfo{volume}{158}}, \bibinfo{pages}{113573}
  (\bibinfo{year}{2020}).

\bibitem[{\citenamefont{Almahdi and Yang}(2019)}]{almahdi2019constrained}
\bibinfo{author}{\bibfnamefont{S.}~\bibnamefont{Almahdi}} \bibnamefont{and}
  \bibinfo{author}{\bibfnamefont{S.~Y.} \bibnamefont{Yang}}, \bibinfo{title}{A
  constrained portfolio trading system using particle swarm algorithm and
  recurrent reinforcement learning}, \emph{\bibinfo{journal}{Expert Systems
  with Applications}} \textbf{\bibinfo{volume}{130}}, \bibinfo{pages}{145}
  (\bibinfo{year}{2019}).

\bibitem[{\citenamefont{Markowitz}(1952)}]{markowitz1952harry}
\bibinfo{author}{\bibfnamefont{H.}~\bibnamefont{Markowitz}},
  \bibinfo{title}{Harry m. markowitz}, \emph{\bibinfo{journal}{Portfolio
  selection, Journal of Finance}} \textbf{\bibinfo{volume}{7}},
  \bibinfo{pages}{77} (\bibinfo{year}{1952}).

\bibitem[{\citenamefont{Dabney et~al.}(2018)\citenamefont{Dabney, Rowland,
  Bellemare, and Munos}}]{dabney2018distributional}
\bibinfo{author}{\bibfnamefont{W.}~\bibnamefont{Dabney}},
  \bibinfo{author}{\bibfnamefont{M.}~\bibnamefont{Rowland}},
  \bibinfo{author}{\bibfnamefont{M.}~\bibnamefont{Bellemare}},
  \bibnamefont{and} \bibinfo{author}{\bibfnamefont{R.}~\bibnamefont{Munos}}, in
  \emph{\bibinfo{booktitle}{Proceedings of the AAAI Conference on Artificial
  Intelligence}} (\bibinfo{year}{2018}), vol.~\bibinfo{volume}{32}.

\bibitem[{\citenamefont{Betancourt and Chen}(2021)}]{betancourt2021deep}
\bibinfo{author}{\bibfnamefont{C.}~\bibnamefont{Betancourt}} \bibnamefont{and}
  \bibinfo{author}{\bibfnamefont{W.-H.} \bibnamefont{Chen}},
  \bibinfo{title}{Deep reinforcement learning for portfolio management of
  markets with a dynamic number of assets}, \emph{\bibinfo{journal}{Expert
  Systems with Applications}} \textbf{\bibinfo{volume}{164}},
  \bibinfo{pages}{114002} (\bibinfo{year}{2021}).

\bibitem[{\citenamefont{Horgan et~al.}(2018)\citenamefont{Horgan, Quan, Budden,
  Barth-Maron, Hessel, Van~Hasselt, and Silver}}]{horgan2018distributed}
\bibinfo{author}{\bibfnamefont{D.}~\bibnamefont{Horgan}},
  \bibinfo{author}{\bibfnamefont{J.}~\bibnamefont{Quan}},
  \bibinfo{author}{\bibfnamefont{D.}~\bibnamefont{Budden}},
  \bibinfo{author}{\bibfnamefont{G.}~\bibnamefont{Barth-Maron}},
  \bibinfo{author}{\bibfnamefont{M.}~\bibnamefont{Hessel}},
  \bibinfo{author}{\bibfnamefont{H.}~\bibnamefont{Van~Hasselt}},
  \bibnamefont{and} \bibinfo{author}{\bibfnamefont{D.}~\bibnamefont{Silver}},
  \bibinfo{title}{Distributed prioritized experience replay},
  \emph{\bibinfo{journal}{arXiv preprint arXiv:1803.00933}}
  (\bibinfo{year}{2018}).

\bibitem[{\citenamefont{Alimoradi and Kashan}(2018)}]{alimoradi2018league}
\bibinfo{author}{\bibfnamefont{M.~R.} \bibnamefont{Alimoradi}}
  \bibnamefont{and} \bibinfo{author}{\bibfnamefont{A.~H.}
  \bibnamefont{Kashan}}, \bibinfo{title}{A league championship algorithm
  equipped with network structure and backward q-learning for extracting stock
  trading rules}, \emph{\bibinfo{journal}{Applied soft computing}}
  \textbf{\bibinfo{volume}{68}}, \bibinfo{pages}{478} (\bibinfo{year}{2018}).

\bibitem[{\citenamefont{Bu and Cho}(2018)}]{bu2018learning}
\bibinfo{author}{\bibfnamefont{S.-J.} \bibnamefont{Bu}} \bibnamefont{and}
  \bibinfo{author}{\bibfnamefont{S.-B.} \bibnamefont{Cho}}, in
  \emph{\bibinfo{booktitle}{International Conference on Intelligent Data
  Engineering and Automated Learning}} (\bibinfo{organization}{Springer},
  \bibinfo{year}{2018}),  \bibinfo{pages}{468--480}.

\bibitem[{\citenamefont{Van~Hasselt et~al.}(2016)\citenamefont{Van~Hasselt,
  Guez, and Silver}}]{van2016deep}
\bibinfo{author}{\bibfnamefont{H.}~\bibnamefont{Van~Hasselt}},
  \bibinfo{author}{\bibfnamefont{A.}~\bibnamefont{Guez}}, \bibnamefont{and}
  \bibinfo{author}{\bibfnamefont{D.}~\bibnamefont{Silver}}, in
  \emph{\bibinfo{booktitle}{Proceedings of the AAAI conference on artificial
  intelligence}} (\bibinfo{year}{2016}), vol.~\bibinfo{volume}{30}.

\bibitem[{\citenamefont{Chen and Gao}(2019)}]{chen2019application}
\bibinfo{author}{\bibfnamefont{L.}~\bibnamefont{Chen}} \bibnamefont{and}
  \bibinfo{author}{\bibfnamefont{Q.}~\bibnamefont{Gao}}, in
  \emph{\bibinfo{booktitle}{2019 IEEE 10th International Conference on Software
  Engineering and Service Science (ICSESS)}} (\bibinfo{organization}{IEEE},
  \bibinfo{year}{2019}),  \bibinfo{pages}{29--33}.

\bibitem[{\citenamefont{Huang}(2018)}]{huang2018financial}
\bibinfo{author}{\bibfnamefont{C.~Y.} \bibnamefont{Huang}},
  \bibinfo{title}{Financial trading as a game: A deep reinforcement learning
  approach}, \emph{\bibinfo{journal}{arXiv preprint arXiv:1807.02787}}
  (\bibinfo{year}{2018}).

\bibitem[{\citenamefont{Lucarelli and Borrotti}(2020)}]{lucarelli2020deep}
\bibinfo{author}{\bibfnamefont{G.}~\bibnamefont{Lucarelli}} \bibnamefont{and}
  \bibinfo{author}{\bibfnamefont{M.}~\bibnamefont{Borrotti}}, \bibinfo{title}{A
  deep q-learning portfolio management framework for the cryptocurrency
  market}, \emph{\bibinfo{journal}{Neural Computing and Applications}}
  \textbf{\bibinfo{volume}{32}}, \bibinfo{pages}{17229} (\bibinfo{year}{2020}).

\bibitem[{\citenamefont{Jeong and Kim}(2019)}]{jeong2019improving}
\bibinfo{author}{\bibfnamefont{G.}~\bibnamefont{Jeong}} \bibnamefont{and}
  \bibinfo{author}{\bibfnamefont{H.~Y.} \bibnamefont{Kim}},
  \bibinfo{title}{Improving financial trading decisions using deep q-learning:
  Predicting the number of shares, action strategies, and transfer learning},
  \emph{\bibinfo{journal}{Expert Systems with Applications}}
  \textbf{\bibinfo{volume}{117}}, \bibinfo{pages}{125} (\bibinfo{year}{2019}).

\bibitem[{\citenamefont{Zhang et~al.}(2020)\citenamefont{Zhang, Zohren, and
  Roberts}}]{zhang2020deep}
\bibinfo{author}{\bibfnamefont{Z.}~\bibnamefont{Zhang}},
  \bibinfo{author}{\bibfnamefont{S.}~\bibnamefont{Zohren}}, \bibnamefont{and}
  \bibinfo{author}{\bibfnamefont{S.}~\bibnamefont{Roberts}},
  \bibinfo{title}{Deep reinforcement learning for trading},
  \emph{\bibinfo{journal}{The Journal of Financial Data Science}}
  \textbf{\bibinfo{volume}{2}}, \bibinfo{pages}{25} (\bibinfo{year}{2020}).

\bibitem[{\citenamefont{Deng et~al.}(2016)\citenamefont{Deng, Bao, Kong, Ren,
  and Dai}}]{deng2016deep}
\bibinfo{author}{\bibfnamefont{Y.}~\bibnamefont{Deng}},
  \bibinfo{author}{\bibfnamefont{F.}~\bibnamefont{Bao}},
  \bibinfo{author}{\bibfnamefont{Y.}~\bibnamefont{Kong}},
  \bibinfo{author}{\bibfnamefont{Z.}~\bibnamefont{Ren}}, \bibnamefont{and}
  \bibinfo{author}{\bibfnamefont{Q.}~\bibnamefont{Dai}}, \bibinfo{title}{Deep
  direct reinforcement learning for financial signal representation and
  trading}, \emph{\bibinfo{journal}{IEEE transactions on neural networks and
  learning systems}} \textbf{\bibinfo{volume}{28}}, \bibinfo{pages}{653}
  (\bibinfo{year}{2016}).

\bibitem[{\citenamefont{Wang et~al.}(2016)\citenamefont{Wang, Schaul, Hessel,
  Hasselt, Lanctot, and Freitas}}]{wang2016dueling}
\bibinfo{author}{\bibfnamefont{Z.}~\bibnamefont{Wang}},
  \bibinfo{author}{\bibfnamefont{T.}~\bibnamefont{Schaul}},
  \bibinfo{author}{\bibfnamefont{M.}~\bibnamefont{Hessel}},
  \bibinfo{author}{\bibfnamefont{H.}~\bibnamefont{Hasselt}},
  \bibinfo{author}{\bibfnamefont{M.}~\bibnamefont{Lanctot}}, \bibnamefont{and}
  \bibinfo{author}{\bibfnamefont{N.}~\bibnamefont{Freitas}}, in
  \emph{\bibinfo{booktitle}{International conference on machine learning}}
  (\bibinfo{organization}{PMLR}, \bibinfo{year}{2016}),
  \bibinfo{pages}{1995--2003}.

\bibitem[{\citenamefont{Wu et~al.}(2020)\citenamefont{Wu, Chen, Wang, Troiano,
  Loia, and Fujita}}]{wu2020adaptive}
\bibinfo{author}{\bibfnamefont{X.}~\bibnamefont{Wu}},
  \bibinfo{author}{\bibfnamefont{H.}~\bibnamefont{Chen}},
  \bibinfo{author}{\bibfnamefont{J.}~\bibnamefont{Wang}},
  \bibinfo{author}{\bibfnamefont{L.}~\bibnamefont{Troiano}},
  \bibinfo{author}{\bibfnamefont{V.}~\bibnamefont{Loia}}, \bibnamefont{and}
  \bibinfo{author}{\bibfnamefont{H.}~\bibnamefont{Fujita}},
  \bibinfo{title}{Adaptive stock trading strategies with deep reinforcement
  learning methods}, \emph{\bibinfo{journal}{Information Sciences}}
  \textbf{\bibinfo{volume}{538}}, \bibinfo{pages}{142} (\bibinfo{year}{2020}).

\bibitem[{\citenamefont{Vezeris et~al.}(2019)\citenamefont{Vezeris, Karkanis,
  and Kyrgos}}]{vezeris2019adturtle}
\bibinfo{author}{\bibfnamefont{D.}~\bibnamefont{Vezeris}},
  \bibinfo{author}{\bibfnamefont{I.}~\bibnamefont{Karkanis}}, \bibnamefont{and}
  \bibinfo{author}{\bibfnamefont{T.}~\bibnamefont{Kyrgos}},
  \bibinfo{title}{Adturtle: An advanced turtle trading system},
  \emph{\bibinfo{journal}{Journal of Risk and Financial Management}}
  \textbf{\bibinfo{volume}{12}}, \bibinfo{pages}{96} (\bibinfo{year}{2019}).

\bibitem[{\citenamefont{Liang et~al.}(2018)\citenamefont{Liang, Chen, Zhu,
  Jiang, and Li}}]{liang2018adversarial}
\bibinfo{author}{\bibfnamefont{Z.}~\bibnamefont{Liang}},
  \bibinfo{author}{\bibfnamefont{H.}~\bibnamefont{Chen}},
  \bibinfo{author}{\bibfnamefont{J.}~\bibnamefont{Zhu}},
  \bibinfo{author}{\bibfnamefont{K.}~\bibnamefont{Jiang}}, \bibnamefont{and}
  \bibinfo{author}{\bibfnamefont{Y.}~\bibnamefont{Li}},
  \bibinfo{title}{Adversarial deep reinforcement learning in portfolio
  management}, \emph{\bibinfo{journal}{arXiv preprint arXiv:1808.09940}}
  (\bibinfo{year}{2018}).

\bibitem[{\citenamefont{Jiang et~al.}(2017)\citenamefont{Jiang, Xu, and
  Liang}}]{jiang2017deep}
\bibinfo{author}{\bibfnamefont{Z.}~\bibnamefont{Jiang}},
  \bibinfo{author}{\bibfnamefont{D.}~\bibnamefont{Xu}}, \bibnamefont{and}
  \bibinfo{author}{\bibfnamefont{J.}~\bibnamefont{Liang}}, \bibinfo{title}{A
  deep reinforcement learning framework for the financial portfolio management
  problem}, \emph{\bibinfo{journal}{arXiv preprint arXiv:1706.10059}}
  (\bibinfo{year}{2017}).

\bibitem[{\citenamefont{Pendharkar and Cusatis}(2018)}]{pendharkar2018trading}
\bibinfo{author}{\bibfnamefont{P.~C.} \bibnamefont{Pendharkar}}
  \bibnamefont{and} \bibinfo{author}{\bibfnamefont{P.}~\bibnamefont{Cusatis}},
  \bibinfo{title}{Trading financial indices with reinforcement learning
  agents}, \emph{\bibinfo{journal}{Expert Systems with Applications}}
  \textbf{\bibinfo{volume}{103}}, \bibinfo{pages}{1} (\bibinfo{year}{2018}).

\bibitem[{\citenamefont{Schnaubelt}(2022)}]{schnaubelt2022deep}
\bibinfo{author}{\bibfnamefont{M.}~\bibnamefont{Schnaubelt}},
  \bibinfo{title}{Deep reinforcement learning for the optimal placement of
  cryptocurrency limit orders}, \emph{\bibinfo{journal}{European Journal of
  Operational Research}} \textbf{\bibinfo{volume}{296}}, \bibinfo{pages}{993}
  (\bibinfo{year}{2022}).

\bibitem[{\citenamefont{Bertoluzzo and Corazza}(2012)}]{bertoluzzo2012testing}
\bibinfo{author}{\bibfnamefont{F.}~\bibnamefont{Bertoluzzo}} \bibnamefont{and}
  \bibinfo{author}{\bibfnamefont{M.}~\bibnamefont{Corazza}},
  \bibinfo{title}{Testing different reinforcement learning configurations for
  financial trading: Introduction and applications},
  \emph{\bibinfo{journal}{Procedia Economics and Finance}}
  \textbf{\bibinfo{volume}{3}}, \bibinfo{pages}{68} (\bibinfo{year}{2012}).

\bibitem[{\citenamefont{Taghian et~al.}(2022)\citenamefont{Taghian, Asadi, and
  Safabakhsh}}]{taghian2022learning}
\bibinfo{author}{\bibfnamefont{M.}~\bibnamefont{Taghian}},
  \bibinfo{author}{\bibfnamefont{A.}~\bibnamefont{Asadi}}, \bibnamefont{and}
  \bibinfo{author}{\bibfnamefont{R.}~\bibnamefont{Safabakhsh}},
  \bibinfo{title}{Learning financial asset-specific trading rules via deep
  reinforcement learning}, \emph{\bibinfo{journal}{Expert Systems with
  Applications}} \textbf{\bibinfo{volume}{195}}, \bibinfo{pages}{116523}
  (\bibinfo{year}{2022}).

\bibitem[{\citenamefont{Bellemare et~al.}(2017)\citenamefont{Bellemare, Dabney,
  and Munos}}]{bellemare2017distributional}
\bibinfo{author}{\bibfnamefont{M.~G.} \bibnamefont{Bellemare}},
  \bibinfo{author}{\bibfnamefont{W.}~\bibnamefont{Dabney}}, \bibnamefont{and}
  \bibinfo{author}{\bibfnamefont{R.}~\bibnamefont{Munos}}, in
  \emph{\bibinfo{booktitle}{International Conference on Machine Learning}}
  (\bibinfo{organization}{PMLR}, \bibinfo{year}{2017}),
  \bibinfo{pages}{449--458}.

\bibitem[{\citenamefont{Lillicrap et~al.}(2015)\citenamefont{Lillicrap, Hunt,
  Pritzel, Heess, Erez, Tassa, Silver, and Wierstra}}]{lillicrap2015continuous}
\bibinfo{author}{\bibfnamefont{T.~P.} \bibnamefont{Lillicrap}},
  \bibinfo{author}{\bibfnamefont{J.~J.} \bibnamefont{Hunt}},
  \bibinfo{author}{\bibfnamefont{A.}~\bibnamefont{Pritzel}},
  \bibinfo{author}{\bibfnamefont{N.}~\bibnamefont{Heess}},
  \bibinfo{author}{\bibfnamefont{T.}~\bibnamefont{Erez}},
  \bibinfo{author}{\bibfnamefont{Y.}~\bibnamefont{Tassa}},
  \bibinfo{author}{\bibfnamefont{D.}~\bibnamefont{Silver}}, \bibnamefont{and}
  \bibinfo{author}{\bibfnamefont{D.}~\bibnamefont{Wierstra}},
  \bibinfo{title}{Continuous control with deep reinforcement learning},
  \emph{\bibinfo{journal}{arXiv preprint arXiv:1509.02971}}
  (\bibinfo{year}{2015}).

\bibitem[{\citenamefont{Fujimoto et~al.}(2018)\citenamefont{Fujimoto, Hoof, and
  Meger}}]{fujimoto2018addressing}
\bibinfo{author}{\bibfnamefont{S.}~\bibnamefont{Fujimoto}},
  \bibinfo{author}{\bibfnamefont{H.}~\bibnamefont{Hoof}}, \bibnamefont{and}
  \bibinfo{author}{\bibfnamefont{D.}~\bibnamefont{Meger}}, in
  \emph{\bibinfo{booktitle}{International conference on machine learning}}
  (\bibinfo{organization}{PMLR}, \bibinfo{year}{2018}),
  \bibinfo{pages}{1587--1596}.

\bibitem[{\citenamefont{Carta et~al.}(2021)\citenamefont{Carta, Corriga,
  Ferreira, Podda, and Recupero}}]{carta2021multi}
\bibinfo{author}{\bibfnamefont{S.}~\bibnamefont{Carta}},
  \bibinfo{author}{\bibfnamefont{A.}~\bibnamefont{Corriga}},
  \bibinfo{author}{\bibfnamefont{A.}~\bibnamefont{Ferreira}},
  \bibinfo{author}{\bibfnamefont{A.~S.} \bibnamefont{Podda}}, \bibnamefont{and}
  \bibinfo{author}{\bibfnamefont{D.~R.} \bibnamefont{Recupero}},
  \bibinfo{title}{A multi-layer and multi-ensemble stock trader using deep
  learning and deep reinforcement learning}, \emph{\bibinfo{journal}{Applied
  Intelligence}} \textbf{\bibinfo{volume}{51}}, \bibinfo{pages}{889}
  (\bibinfo{year}{2021}).

\bibitem[{\citenamefont{Barra et~al.}(2020)\citenamefont{Barra, Carta, Corriga,
  Podda, and Recupero}}]{barra2020deep}
\bibinfo{author}{\bibfnamefont{S.}~\bibnamefont{Barra}},
  \bibinfo{author}{\bibfnamefont{S.~M.} \bibnamefont{Carta}},
  \bibinfo{author}{\bibfnamefont{A.}~\bibnamefont{Corriga}},
  \bibinfo{author}{\bibfnamefont{A.~S.} \bibnamefont{Podda}}, \bibnamefont{and}
  \bibinfo{author}{\bibfnamefont{D.~R.} \bibnamefont{Recupero}},
  \bibinfo{title}{Deep learning and time series-to-image encoding for financial
  forecasting}, \emph{\bibinfo{journal}{IEEE/CAA Journal of Automatica Sinica}}
  \textbf{\bibinfo{volume}{7}}, \bibinfo{pages}{683} (\bibinfo{year}{2020}).

\bibitem[{\citenamefont{Th{\'e}ate and Ernst}(2021)}]{theate2021application}
\bibinfo{author}{\bibfnamefont{T.}~\bibnamefont{Th{\'e}ate}} \bibnamefont{and}
  \bibinfo{author}{\bibfnamefont{D.}~\bibnamefont{Ernst}}, \bibinfo{title}{An
  application of deep reinforcement learning to algorithmic trading},
  \emph{\bibinfo{journal}{Expert Systems with Applications}}
  \textbf{\bibinfo{volume}{173}}, \bibinfo{pages}{114632}
  (\bibinfo{year}{2021}).

\bibitem[{\citenamefont{Rockafellar et~al.}(2000)\citenamefont{Rockafellar,
  Uryasev et~al.}}]{rockafellar2000optimization}
\bibinfo{author}{\bibfnamefont{R.~T.} \bibnamefont{Rockafellar}},
  \bibinfo{author}{\bibfnamefont{S.}~\bibnamefont{Uryasev}},
  \bibnamefont{et~al.}, \bibinfo{title}{Optimization of conditional
  value-at-risk}, \emph{\bibinfo{journal}{Journal of risk}}
  \textbf{\bibinfo{volume}{2}}, \bibinfo{pages}{21} (\bibinfo{year}{2000}).

\end{thebibliography}

\end{document}